\documentclass[reprint]{JASA}

\usepackage{algpseudocode}
\usepackage{xcolor, soul}
\usepackage{multirow}
\sethlcolor{yellow}
\usepackage{ulem}

\newif\ifnotes
\makeatletter
\newcommand{\note}[1]{\@bsphack\ifnotes{#1}\fi\@esphack}
\makeatother

\notesfalse 

\def\alphavect{\mbox{\boldmath $\alpha$}}

\begin{document}

\title[JASA/Mean Absorption Estimation]{Mean Absorption Estimation from Room Impulse Responses using Virtually-Supervised Learning}
\author{Cédric Foy}
\email{cedric.foy@cerema.fr}
\affiliation{UMRAE, Cerema, Univ. Gustave Eiffel, Ifsttar, Strasbourg, 67035, France}

\author{Antoine Deleforge}
\email{antoine.deleforge@inria.fr}
\affiliation{Universit\'e de Lorraine, CNRS, Inria, LORIA, F-54000 Nancy, France}

\author{Diego Di Carlo}
\email{diego.di-carlo@inria.fr}
\affiliation{Univ Rennes, Inria, CNRS, IRISA, France}


\date{\today} 

\begin{abstract}
In the context of building acoustics and the acoustic diagnosis of an existing room, this paper introduces and investigates a new approach to estimate mean absorption coefficients solely from a room impulse response (RIR). This inverse problem is tackled via \textit{virtually-supervised learning}, namely, the RIR-to-absorption mapping is implicitly learned by regression on a simulated dataset using artificial neural networks. We focus on simple models based on well-understood architectures. The critical choices of geometric, acoustic and simulation parameters used to train the models are extensively discussed and studied, while keeping in mind conditions that are representative of the field of building acoustics. Estimation errors from the learned neural models are compared to those obtained with classical formulas that \textcolor{black}{require knowledge of the room's geometry and reverberation times}. Extensive comparisons made on a variety of simulated test sets highlight different conditions under which the learned models can overcome the well-known limitations of the diffuse sound field hypothesis underlying these formulas. Results obtained on real RIRs measured in an acoustically configurable room show that at 1~kHz and above, \textcolor{black}{the proposed approach performs comparably to classical models when reverberation times can be reliably estimated, and continues to work even when they cannot.}
\end{abstract}


\maketitle


\section{\label{sec:intro} Introduction}
When sound propagates in a room, its reflections on the walls, ceiling, floor and other surfaces lead to the well known phenomenon of reverberation. When the reverberation level is too high, it can be a major source of nuisance for the room's users.
\note{\sout{To improve the listening quality of a room, the main parameters}}
\textcolor{black}{To alleviate this, some of the main parameters} an acoustician can act on are the \textit{absorption coefficients} of the room surfaces, namely, the proportion of sound energy that the surfaces' materials do not reflect. These are generally frequency-dependent and are typically expressed within octave bands, $b\in\mathcal{F}=\{.125,.25,.5,1,2,4\}$~kHz in room acoustics standards. To obtain the acoustic diagnosis of a room and deduce a renovation plan, acousticians need to know the absorption coefficients $\alpha_i(b)$ of each individual surface $i$ in the room. This is typically done through a manual iterative process where acoustic simulators are tuned to match \textit{in situ} measurements while taking into account the room's geometry and the properties of known materials, as measured in \textcolor{black}{laboratories}.

Among \textit{in situ} measurements used in practice, room impulse responses (RIR) are rich signals that capture the acoustic signature of the room via the shape of their decay, their echo density over time
or the timings of their early echoes.
While the \textit{forward} physical process from acoustic parameters to RIRs is well understood, as illustrated by the existence of many reasonably accurate and efficient RIR simulators \cite{habets2006room,schimmel2009fast,scheibler2018pyroomacoustics}, the \textit{inverse problem} of retrieving the absorption coefficients of surfaces solely from a RIR is much more challenging and is the focus of this article.
\note{\sout{To obtain the acoustic diagnosis of a room and deduce a renovation plan, one would ideally need to know the absorption coefficients $\alpha_i(b)$ of each individual surface $i$ in the room. However, no method is currently known to reliably estimate these quantities \textit{in situ}. In practice, the \textit{mean absorption coefficients} are estimated instead. These are defined as}}
\textcolor{black}{We consider the simple but common case of a \textit{shoebox} (cuboid) room with a different material on each of the 6 surfaces. Even in this case, recovering the absorption coefficients of all surfaces from a single RIR without any knowledge on the source, receiver or wall positions is out of reach, due to inherent ambiguities of the problem such as permutations between the different surfaces. To alleviate this issue, this work focuses on estimating the area-weighted mean absorption coefficients:}
\begin{equation}
\label{eq:alpha}
\bar{\alpha}(b) = \frac{\sum_i\alpha_i(b)S_i}{\sum_i S_i}\in[0,1]
\end{equation}
where $S_i$ denotes the area of surface $i$ in $\textrm{m}^2$.
\note{\sout{Currently, the standard method to estimate $\bar{\alpha}(b)$ \textit{in situ} is to calculate the room's reverberation time $RT(b)$ by Schroeder integration of a measured room impulse response (RIR) \mbox{\cite{Schroeder:65}} and to use it in the celebrated Sabine's law or its more precise variant from Eyring, based on reverberation theory \mbox{\cite{Kuttruff:09}}: }}
\note{\sout{where $V$ denotes the room's volume and $S=\sum_i S_i$ its total surface. This standard methodology suffers from known limitations and approximations. First, it requires to know the room's geometry by means of $S$ and $V$. Second, it requires an accurate estimation of the reverberation time, which may not be available when a measured RIR features an insufficient or non linear decay of its Schroeder curve \mbox{\cite{Schroeder:65, Xiang1995}}. Last, reverberation theory strongly relies on the assumption of a diffuse sound field, homogeneous at all points and in all directions of the room. This assumption holds well in approximately cubic rooms, with low and homogeneous absorption coefficients on all surfaces, and with a source and receiver placed far from reflective surfaces. When these assumptions are not met, reverberation theory estimates severely degrade in practice.}}
\note{\sout{To alleviate these limits, an attractive idea is to estimate $\bar{\alpha}$ directly from full RIRs rather than merely reverberation times.}}
Note that this quantity is treated here as a purely analytical parameter that globally summarizes the acoustic properties of all surfaces in the room. In acoustics, it is traditionally \textcolor{black}{used} under the hypothesis of a \textcolor{black}{\textit{diffuse sound field} (DSF) in which} the energy is uniformly distributed in space and flows isotropically \cite{Kuttruff:09,nolan2018wavenumber}. However, in this work, we will also consider its estimation under more general, non-diffuse settings. Choosing this particular quantity as a target will notably allow relevant comparisons to methods based on classical reverberation theory, \textit{i.e.}, by inverting the well-known Sabine and Eyring formulas \cite{Kuttruff:09}, at least under conditions that are close to the DSF regime.
\note{\sout{This assumption holds well in approximately cubic rooms, with low and homogeneous absorption coefficients on all surfaces, and with a source and receiver placed far from reflective surfaces. When these assumptions are not met, reverberation theory estimates severely degrade in practice.}}

We propose to tackle the inverse problem of estimating $\bar{\alphavect}=[\bar{\alpha}(b)]_{b\in\mathcal{F}}\in[0,1]^6$ from a single RIR without any other information on the room using supervised machine learning and in particular non-linear regression.

While artificial neural networks have proven to be a very powerful family of models for non-linear regression in the recent years, a well-known bottleneck is their need for a large number of input-output pairs to be trained.
\note{\sout{Given a set of input-output pairs $\{(x_n,y_n)\}_{n=1}^N\subseteq\mathbb{R}^{D_{\textrm{in}}}\times\mathbb{R}^{D_{\textrm{out}}}$ and a parameterized family of non-linear functions $\{f_\theta\}_{\theta}$, the task is to estimate parameters $\theta$ such that $f_\theta(x_n)$ is as close a possible to $y_n$ for all $n$.}\sout{Over the past decade, deep neural networks have come forward as a powerful family of such functions, in particular due to their remarkable ability to generalize to unseen data. This success was notably made possible by advances in stochastic gradient descent optimization techniques \mbox{\cite{kingma2014adam}} and ever increasing computational capabilities.}}
As of today, since \textit{in situ} estimation of absorption coefficients remains a costly and complex task, sufficiently large and diverse real RIR databases annotated with surface absorption profiles are not available. Hence we propose to make use of \textit{virtually supervised learning}, as introduced in \cite{gaultier2017vast}. The idea is to use the known forward physical model, namely, a room acoustic simulator, to generate a potentially unlimited amount of annotated data to learn the inverse mapping from.
The main contributions of this article are (i) a novel approach to efficiently sample simulated training data that are representative of commonly encountered acoustics in cuboid rooms, which is shown to outperform naive uniform sampling; (ii) an extensive comparative simulation study between estimates based on classical reverberation theory and those obtained from various neural network designs, including their generalizability to unseen data, noise, and various acoustic conditions; and (iii) a comparative study between virtually trained models and classical models on real measured RIRs.

\note{\sout{In simulations, the proposed mean absorption coefficient estimation method is showed to outperform conventional methods, despite not requiring the room's geometry. A reduction of estimation errors on mean absorption coefficients by a factor of up to two is observed on simulated data, in particular in difficult non-homogeneous and noisy conditions. Results on hundreds of real RIRs measured in an acoustically configurable rectangular room show that the approach performs comparably to Eyring above 500 Hz in most conditions, but crucially, continues to work even when good reverberation times are not available. Further studies and extensions of the method to establish its real-world applicability at lower frequencies and in more general acoustic environments are left for future work.}}

\textcolor{black}{Our simulated experiments reveal that neural models can successfully estimate mean absorption coefficients under a wide range of acoustical conditions, with mean absolute errors below 0.05, while not requiring any geometrical information on the room. As expected, in non-DSF settings, they are more accurate than classical models that rely on the DSF hypothesis. On real data that are close to the DSF regime, errors obtained from the proposed learned model are not satisfying below $1~\textrm{kHz}$ but remain under 0.1 in higher octave bands and are comparable to those obtained with classical models. Moreover, in those higher frequencies, it is shown that the neural model continues to yield reliable $\bar{\alpha}(b)$ estimates even in conditions where classical models cannot, as reverberation times cannot be extracted from RIRs due to the lack of sufficient linear decays in Schroeder curves \cite{Schroeder:65}.}

\textcolor{black}{While the observed limitations of classical formulas from reverberation theory outside of the DSF regime are well-known and expected \cite{nolan2018wavenumber}, they still constitute an interesting comparison point as these tools remain widely used today to obtain initial \textit{in situ} acoustical estimates in practice, \textit{e.g.} \cite{prawda2020evaluation}. Further investigation on the real-world applicability of learned models in lower octave bands and their extension to the geometrically-informed estimation of individual absorption profiles are left for future work.}

The remainder of this work is organized as follows. Section \ref{subsec:related} provides an overview of related works. Section \ref{sec:datasets} details the construction of our simulated RIR datasets, examining trade-offs between computational tractability, realism, and representativity. Section \ref{sec:training} presents the neural networks' design and training. Section \ref{sec:results} and \ref{sec:real_results}  contains our extensive comparative experimental study on both simulated and real data. Finally, section \ref{sec:conclusion} concludes and offers leads for future works.

\section{Related works}
\label{subsec:related}

\subsection{Absorption coefficient estimation}
\label{subsec:related_acoustic}
\textcolor{black}{While this article focuses on the intermediate task of estimating area-weighted mean absorption coefficients in a room, the estimation of individual absorption coefficients or more generally the surface impedance of a material is a vast and long-standing research topic, which is briefly reviewed here}.
\textcolor{black}{The most commonly used techniques require an isolated sample of the studied material in a controlled environment. The impedance tube method is one of the most widely used ones \cite{ISO10534,ASTME1050}} and the associated analytical approach is usually that of Chung and Blaser \cite{Chung1980a,Chung1980b} based on the transfer function between two microphones. \textcolor{black}{Alternatively, the \textit{reverberation room} method \cite{ISO354} uses the theory of reverberation and relies on the DSF hypothesis.}

\textcolor{black}{In contrast, this article explores \textit{in situ} estimation. For a recent exhaustive review of this topic, the reader is referred to \cite{Brandao2015l}. Classically, the goal is to separate the direct wave from the reflected wave in an impulse response, with different constraints that depend on the acoustic environment.} Early approaches include \textit{echo-impulse} methods, where the reflected wave is extracted by eliminating the incident wave and parasite wave using temporal windowing or subtraction. Due to the time-frequency uncertainty relation $\Delta t \Delta f \geq 1$ \cite{Garai1993}, a compromise must then be found between the size of the time-domain filters used and the information loss at low frequencies. Also, in order to have a good temporal separation of the waves, the emitted pulse must be narrow, of flat frequency spectrum and repeatable, which is difficult to have in practice \cite{Yuzawa1975,Davies1979,Cramond1984,Garai1993}.

\textcolor{black}{To overcome these limitations, methods based on stationary noise have been proposed. While \cite{Barry1974,Hollin1977l} use white noise, \cite{Aoshima1981} and \cite{Suzuki1995} later proposed a flat spectrum pulse signal stretched in time by filtering. Other excitation signals were then developed to guarantee a better immunity to background noise, such as MLS \cite{Schroeder1979,Rife1989,Stan2002} and Sine Sweep signals \cite{Muller2001,farina2000,farina2007advancements}. To date, the advantages and disadvantages of these signals are still being studied \cite{Torras2010,Guidorzia2015}.}

In parallel, other works focus on the development of analytical models of propagation. In \cite{Ingard1951}, the sound field of an anechoic room is approximated by a set of plane waves. This was later reiterated in \cite{Ando1968} and \cite{Sides1971}. \cite{Allard1985a} introduced the microphonic doublet approach and the specific impedance, which can be related to \textit{surface impedance} using the linearized Euler equation. This approach is only valid if the distance between the microphones is small compared to the wavelength \cite{Allard1985b,Minten1988,Champoux1988a,Champoux1988b}. More finely, the sound field can be modeled by a set of spherical waves, as proposed in \cite{Champoux1988a} based on the analytical model of \cite{Nobile1985} and later in \cite{Li1997}. \textcolor{black}{Finally, approaches based on the principle of acoustical holography, following \cite{tamura1990spatial}, have also been recently investigated \cite{rathsam2015analysis,richard2017estimation,nolan2020estimation}.}
While simple propagation models are easily invertible, more realistic ones are generally not, requiring the use of more complex and approximate numerical solvers, as well as access to precise details on the acoustic environment that are not always available to field acousticians in practice \cite{Brandao2015l}. 

\textcolor{black}{In summary, estimating the absorption coefficients of a material remains a complex task. It hinges on the choice of a number of parameters that are often correlated with each other and hard to precisely control in practice, such as the excitation signal, the source and receiver properties, the environment (free field, anechoic, reverberant), the experimental setup (number and position of sources and microphones, size of the material under study), the chosen propagation model and the post processing. Developing a generic approach to retrieve absorption profiles \textit{in situ} from a unique RIR measurement at an arbitrary location is hence an attractive research avenue for building acoustics.}

\subsection{\label{subsec:related_ml}Machine-learning in acoustics}
Machine learning methodologies have only recently been applied to acoustics. They are still relatively scarce in the field, but have received fast growing interest \cite{bianco2019machine}. While the lack of a large amount of training data is often a limiting factor, this has been alleviated by the use of massive simulations \cite{gaultier2017vast,kim2017generation}, data augmentation \cite{gamper2018blind} or domain adaptation \cite{he2019adaptation}.
Early successful applications of machine learning to acoustics mostly lied in sound source localization \cite{deleforge2014acoustic,deleforge2015co,gaultier2017vast,chakrabarty2017broadband,he2019adaptation,di2019mirage, niu2017source, lefort2017direct} and in acoustic scene and event classification \cite{parsons2000acoustic, deecke2006automated, gradivsek2017predicting,mesaros2017dcase, mesaros2019acoustic}.
\note{\sout{Application domains include robot audition \mbox{\cite{deleforge2014acoustic,he2019adaptation}}, bioacoustics \mbox{\cite{parsons2000acoustic, deecke2006automated, gradivsek2017predicting}} and underwater acoustics \mbox{\cite{niu2017source,lefort2017direct}}.}}
The concept of \textit{acoustic space learning} was introduced in \cite{deleforge2014acoustic} in the context of sound source localization. A large dataset of broadband audio recordings from different (source, receiver) locations in a fixed room was gathered using a motorized binaural head. A supervised non-linear regression model was then trained on this dataset to learn a mapping from audio features to source directions. This approach is however limited by data availability and does not generalize well to different acoustic environments, as showed in \cite{deleforge2015co}. To alleviate this issue, the concept was later extended to \textit{virtual acoustic space learning} \cite{gaultier2017vast,kataria2017hearing}, in which hundreds of thousands of examples are generated using a room acoustic simulator. In the context of sound localization, such virtually-learned models showed some direct albeit limited generalizability to real data in \cite{gaultier2017vast} and in \cite{chakrabarty2017broadband}. In \cite{he2019adaptation}, a domain adaptation technique was proposed to strengthen this generalizability. 

Closer to our application, supervised learning was recently used to estimate the reverberation time \cite{gamper2018blind} or the volume \cite{genovese2019blind} of a room \textit{blindly}, \textit{i.e.}, from the single channel noisy recording of an unknown speech source. Interestingly, these works use a careful combination of real and simulated data for training. Performances are however naturally limited in such blind settings. In a preliminary study \cite{kataria2017hearing}, virtually-supervised learning was used to jointly estimate the mean absorption coefficients of the walls and the 3D position of a broadband noise source from binaural recordings. The room shape, the receiver position and the properties of the floor and ceiling were fixed and known throughout, while the absorption coefficients of walls were supposed frequency-independent and only results on simulated data were reported. Even more recently, a method to estimate the 6 absorption coefficients of the surfaces of a shoebox room in increasing order in a fixed frequency band from an impulse response was proposed, using a fully-connected deep neural network \cite{yu2020room}. \textcolor{black}{The model was both trained and tested on simulated RIR datasets using the image source method, without diffusion or noise, and with absorption coefficients uniformly drawn at random between 0 and 1. Such absorption distribution is however not representative of commonly encountered room acoustics, as will be showed in Section \ref{subsec:represent}. Reported errors were 30\% to 60\% lower than random guessing, but no comparison to known acoustical models and no experiments on real data were carried out.}

\section{Simulated Datasets}
\label{sec:datasets}
The first step of the proposed virtually-supervised approach is to simulate a large number of room impulse responses (RIRs) paired with corresponding mean absorption coefficients $\bar{\alphavect}$ (\ref{eq:alpha}) to train our models. For this, two important trade-offs must to be considered. The first one is between the realism of simulations and their computational demand, and is governed by the choice of a simulator and the tuning of its internal parameters. 
The second one is between the diversity of considered acoustic environments and the amount of representative data needed to train the model. Both trade-offs are discussed in details in sections \ref{subsec:sim} and \ref{subsec:represent}.

\subsection{Realism trade-off}
\label{subsec:sim}
When simulating RIRs, more realism typically implies higher, sometimes prohibitive computational costs. Existing room acoustic simulators can be divided into three categories \cite{habets2006room}. The first category solves the wave equation in discretized space, time and/or frequency domains. These notably include finite element methods \cite{okuzono2014finite}, boundary-element methods \cite{pietrzyk1998computer} or finite-difference time-domain methods \cite{botteldooren1995finite}. While they can in principle simulate any acoustic conditions and geometry to arbitrary precision, their computational time depends on the space discretization steps used, which conditions attainable wavelengths. In the context of building acoustics, which deals with frequencies as high as 5 kHz within large volumes, accurately generating thousands of RIRs is unfeasible with such methods. A second category includes variants of the well-known image source model, originally proposed in \cite{allen1979image}, many times extended, \textit{e.g.}, \cite{peterson1986simulating,borish1984extension,samarasinghe2018spherical}, and implemented in many widely used acoustic simulators, \textit{e.g.}, 
\cite{schimmel2009fast,habets2006room,scheibler2018pyroomacoustics}. This deterministic method allows very efficient implementations, in particular in cuboid rooms, but only models ideal specular reflections on surfaces and hence lacks realism. The last category includes energetic methods based on Monte Carlo sampling, also known as ray-tracing or particle filtering \cite{kulowski1985algorithmic,schroder2011physically,schimmel2009fast}. Like wave-based methods, these approaches can in principle model arbitrary acoustic conditions, and are particularly well-suited to model surface
\note{\sout{diffusion}}
\textcolor{black}{scattering}. However, their computational time and precision depends on the number of rays (or equivalently particles). For such methods to be tractable in the context of room acoustics, the receiver must typically be approximated by a large receptive field in order to aggregate enough rays.
Alternatively, the diffuse-rain method proposed in \cite{schroder2011physically} systematically sends a proportion of diffuse energy to a point receiver at each ray collision, reducing the number of rays needed. In both cases, the timings of rays reaching the receivers are non-deterministic and only reflect acoustical effects in a statistical, energetic sense. 

For this study, we choose a hybrid simulator belonging to the last two categories, referred to as Roomsim and proposed in \cite{schimmel2009fast}. Roomsim combines the image source method to obtain precise timings of specular reflections dominating the early part of the RIR, and the diffuse-rain method to account for stochastic diffuse effects dominating the RIR's tail. The hybrid simulator Roomsim enables frequency-dependent absorption and scattering coefficients and it uses a minimum-phase finite-impulse-response representation of rays reaching the receiver to convert echograms into RIRs. This minimum phase representation is physically motivated by the causality and the fast-decaying properties of resulting signals. \textcolor{black}{A software based on Roomsim is showed to yield remarkably accurate RIRs compared to measured ones in identical conditions in \cite{wabnitz2010room}}. We used the open-source C++/Matlab implementation from the original authors \cite{schimmel2009fast}. As a compromise between accuracy and computational demand, we used a frequency of sampling of 48~kHz, 50,000 rays per simulation for the diffuse-rain method and \textcolor{black}{image sources} up to order 50 for the image-source method. Simulations were run and aggregated along the following 6 octave bands: $b \in\mathcal{F}$. These match those available in most absorption coefficient databases and are commonly used in acoustic regulations. \textcolor{black}{Although its impact is minor, atmospheric attenuation is taken into account for a temperature of 20 degrees Celsius and a relative humidity of 42\% (Roomsim default values).}

We must stress that while lower frequency are perceptually relevant in building acoustics, the \textcolor{black}{energy-based} simulation approach used here is unable to accurately model \textcolor{black}{some of the wave phenomena occurring below the Shroeder's frequency \cite{Schroeder:96} such as room modes \cite[Sec. 5.6]{schroder2011physically}.} This limitation of the current study will be reflected in our real-data experiments, as discussed in section \ref{sec:real_results}.

\subsection{Representativity trade-off}
\label{subsec:represent}
\begin{figure*}[ht!] 
	\centering
	\figline{
		\leftfig{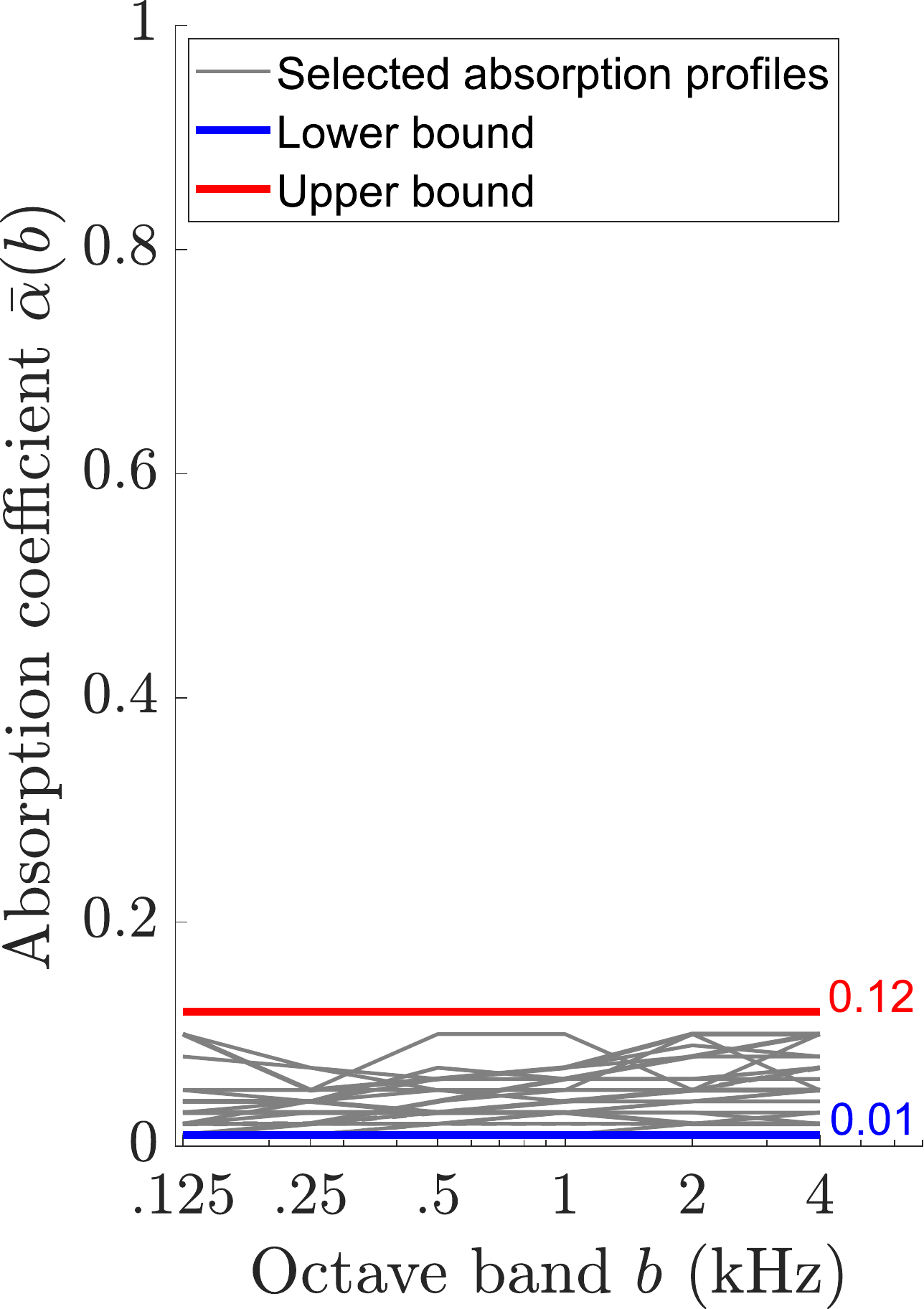}
		{.27\textwidth}
		{(a)}
		\label{fig:profiles_reflective}
		\hfill
		\leftfig{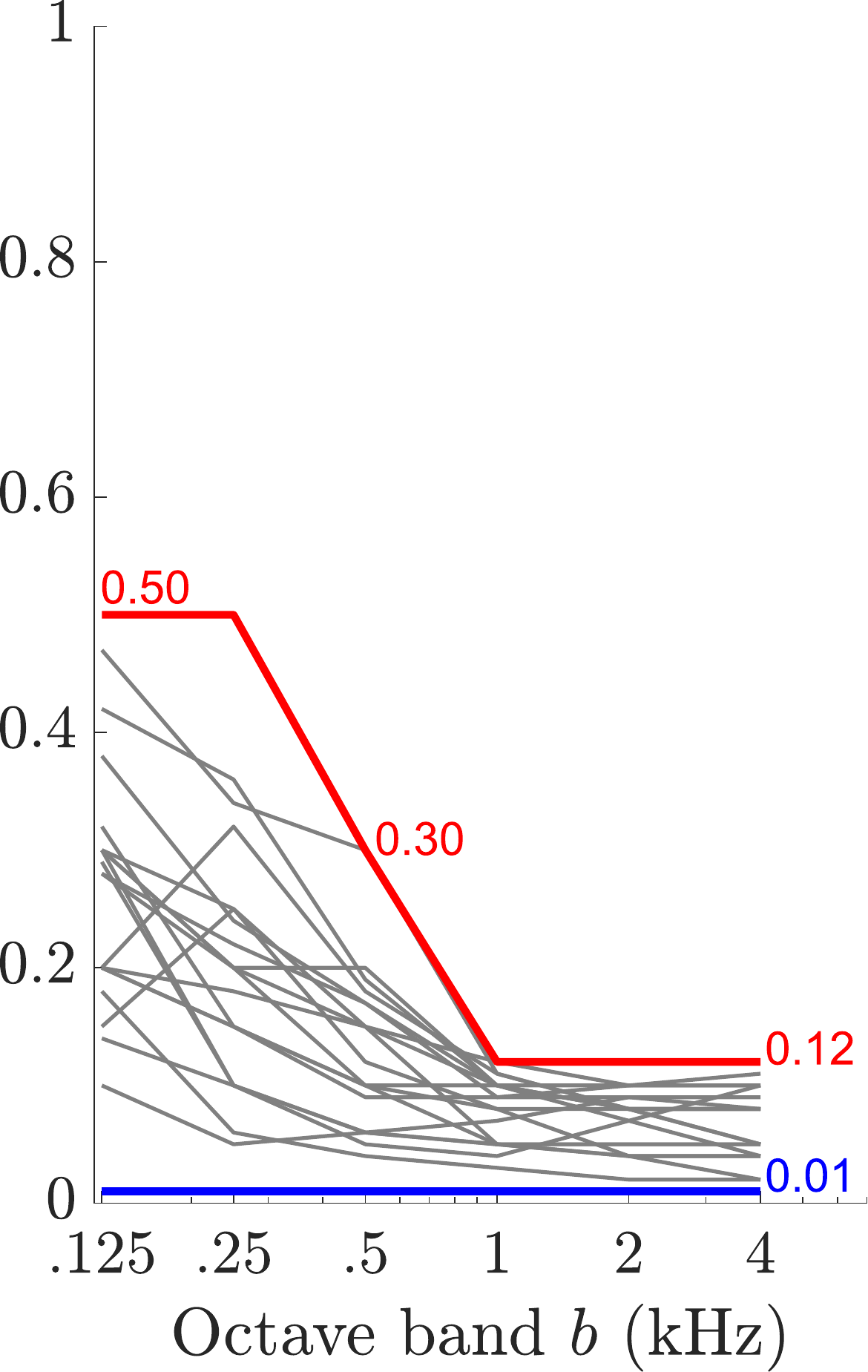}
		{.24\textwidth}
		{(b)}
		\label{fig:profiles_walls}
		\hfill
		\leftfig{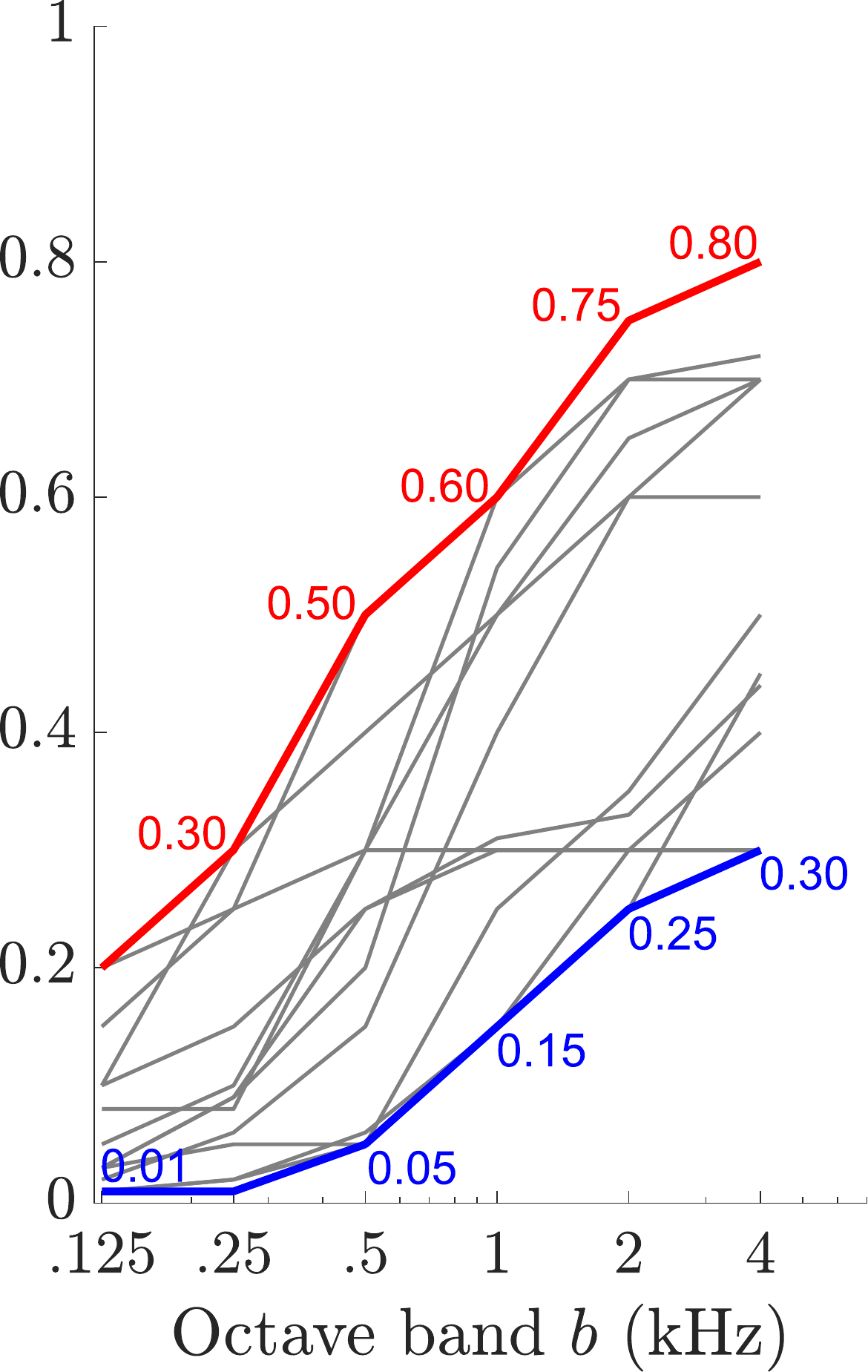}
		{.24\textwidth}
		{(c)}
		\label{fig:profiles_floors}
		\hfill
		\leftfig{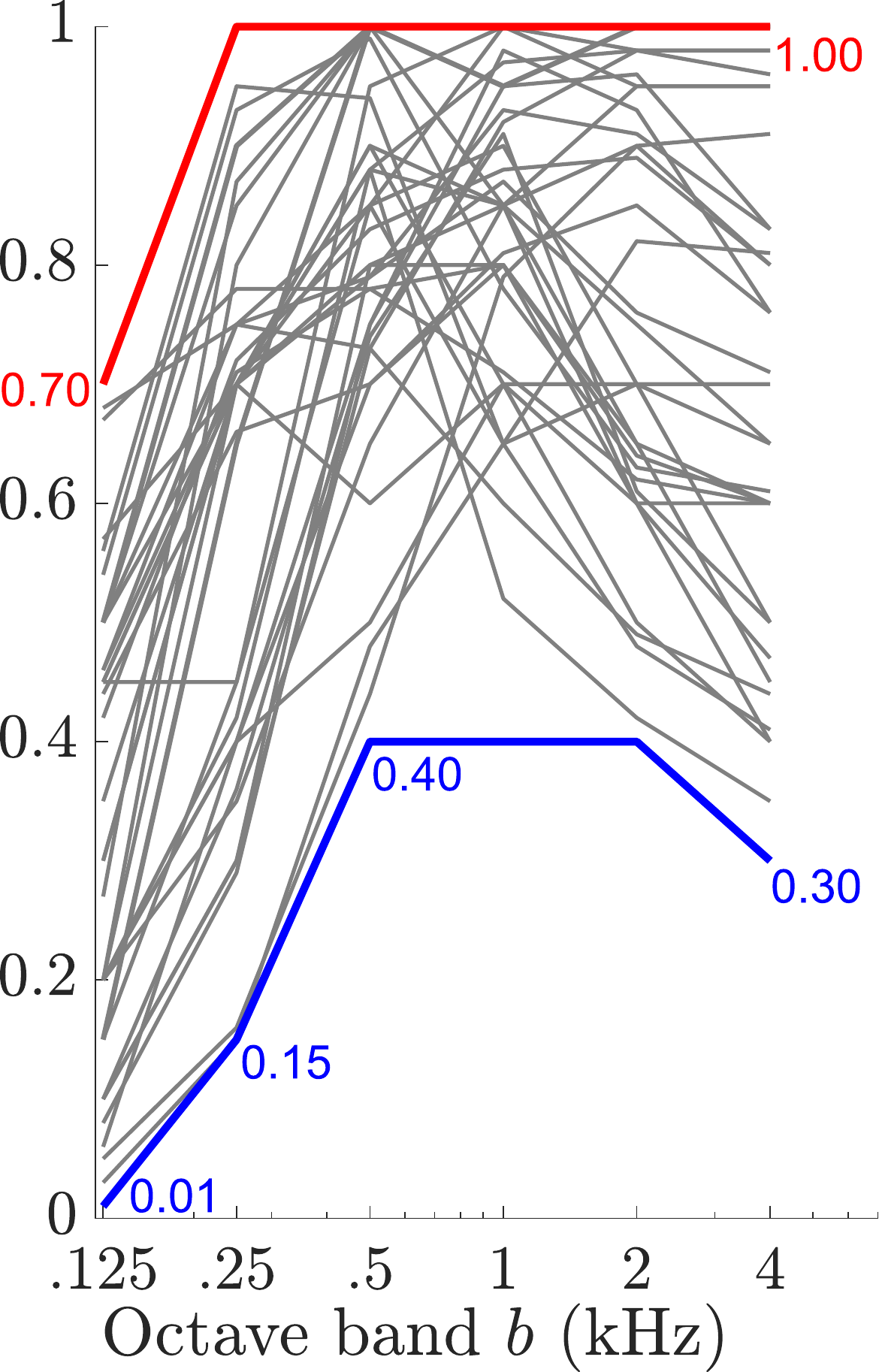}
		{.24\textwidth}
		{(d)}
		\label{fig:profiles_ceilings}
		\hfill
	}
	\caption{\label{fig:profiles} Absorption profiles of 92 commonly encountered reflective, wall, floor and ceiling materials 
		with lower and upper bounds.
		(a) 26 reflective profiles,
		(b) 19 wall profiles,
		(c) 12 floor profiles,
		(d) 35 ceiling profiles.
	}
\end{figure*}

A large diversity in training data is generally desirable to learn a model that generalizes well to many different situations. However, more diversity also implies more data in order to obtain a representative training dataset. Indeed, for a fixed sampling density of a parameterized observation space, the number of required samples grows exponentially in the number of parameters, an effect known as the \textit{curse of dimensionality}. As a mitigating trade-off, we choose in this study to focus on environments that are representative of the field of building acoustics, \textit{e.g.}, offices, schools, restaurants or accommodations. In particular, we exclude very large volumes such as those encountered in churches, tunnels, hangars or swimming pools. Our evaluation will also exclude unusual absorption profiles that are only encountered in highly specialized rooms (\textit{e.g.}, anechoic or semi-anechoic chambers). Fig.~\ref{fig:profiles} shows the absorption profiles of the 92 commonly encountered reflective, wall, floor and ceiling materials that will be considered in this study\footnote{The full lists of materials and associated absorption profiles considered in this study are available here: \url{https://members.loria.fr/ADeleforge/files/jasa2021_supplementary_material.zip}.}. Since most commonly encountered rooms in buildings are cuboids, this study focuses on those rather than dealing with arbitrary complex geometries. This is also motivated by the fact that the image source method is much faster in this setting, as exploited by Roomsim. Finally, we only consider empty rooms. This strong assumption is partially mitigated by the use of the diffuse-rain model. The random sound rays stemming from this Monte Carlo approach can approximate reflections on objects of different sizes, depending on the octave bands/wavelengths considered.

The relevant parameters impacting RIRs can then be divided into a reasonably small set of geometric and acoustic parameters. Geometric parameters include the 3D positions of the source and receiver (both assumed omnidirectional in this study), and the width $L_x$, length $L_y$ and height $L_z$ of the room. The height $L_z$ was drawn uniformly at random between 2.5~m and 4~m and the width $L_x$ and length $L_y$ between 1.5~m and 10~m. \textcolor{black}{The receiver and source positions were drawn uniformly at random in the room for each RIR, while ensuring a minimum distance of 0.5~m to any surface, and 1 m between the two using rejection sampling \cite{ISO3382}}.

\begin{figure*}[ht!]
	\centering
	\figline{
		\leftfig{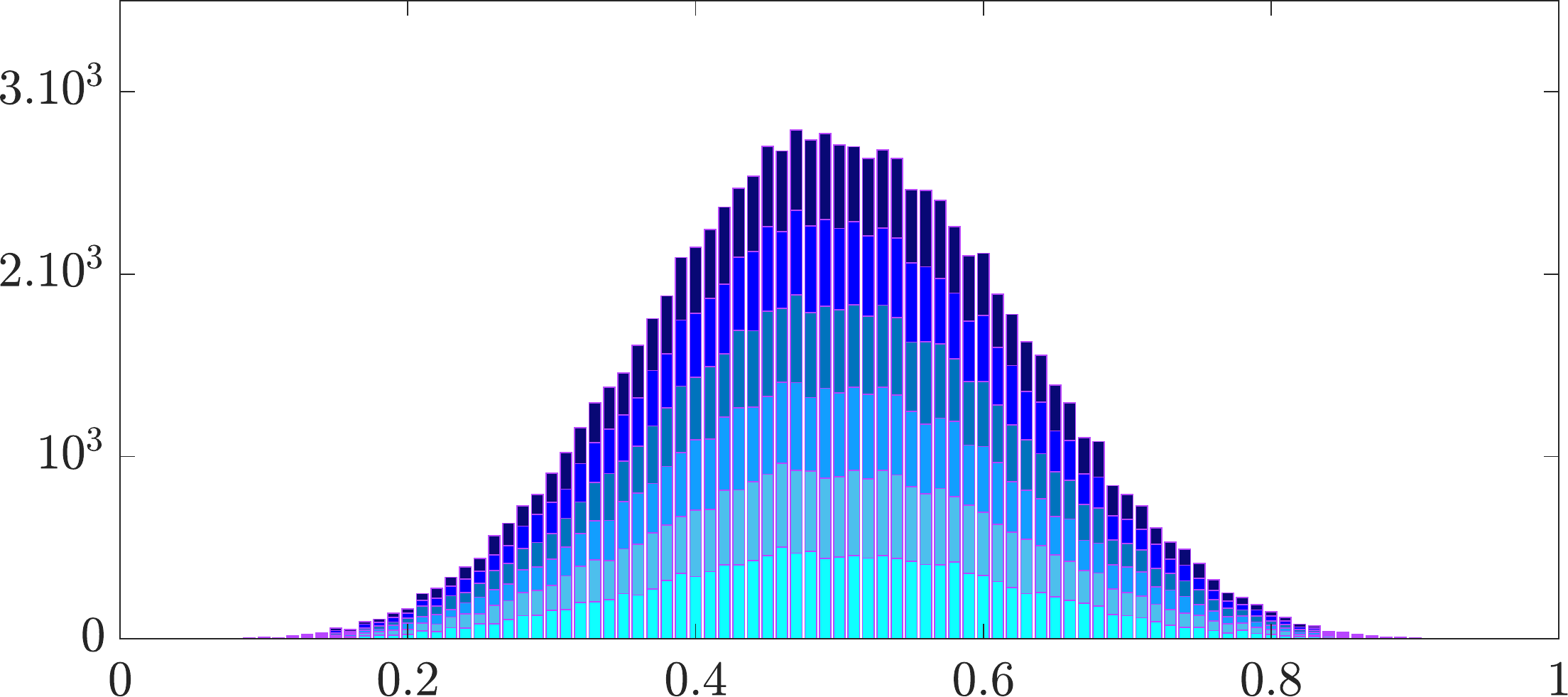}{.49\textwidth}{(a)}\label{fig:hist_alpha_Unif}
		\leftfig{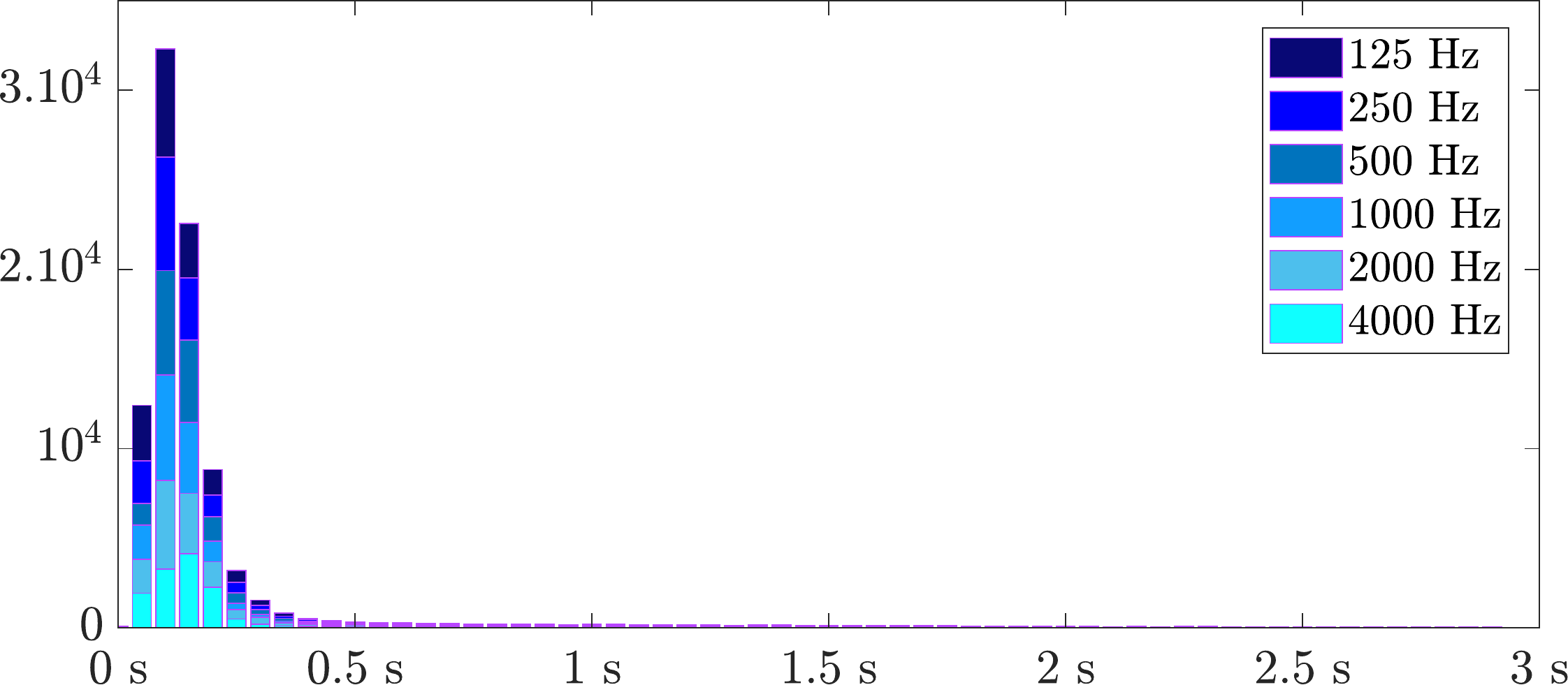}{.49\textwidth}{(b)}\label{fig:hist_RT_Unif}\hfill
	}
	\figline{
		\leftfig{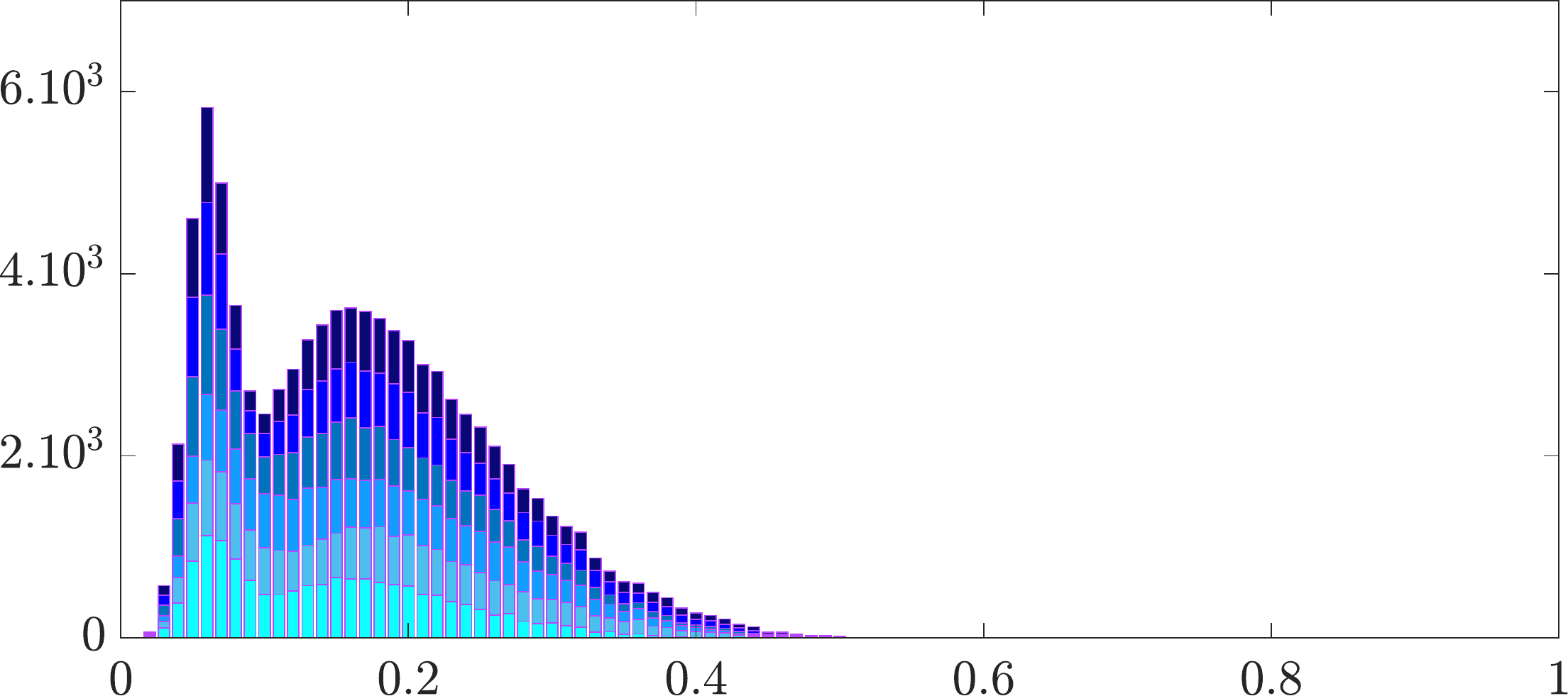}{.49\textwidth}{(c)}\label{fig:hist_alpha_RB}
		\leftfig{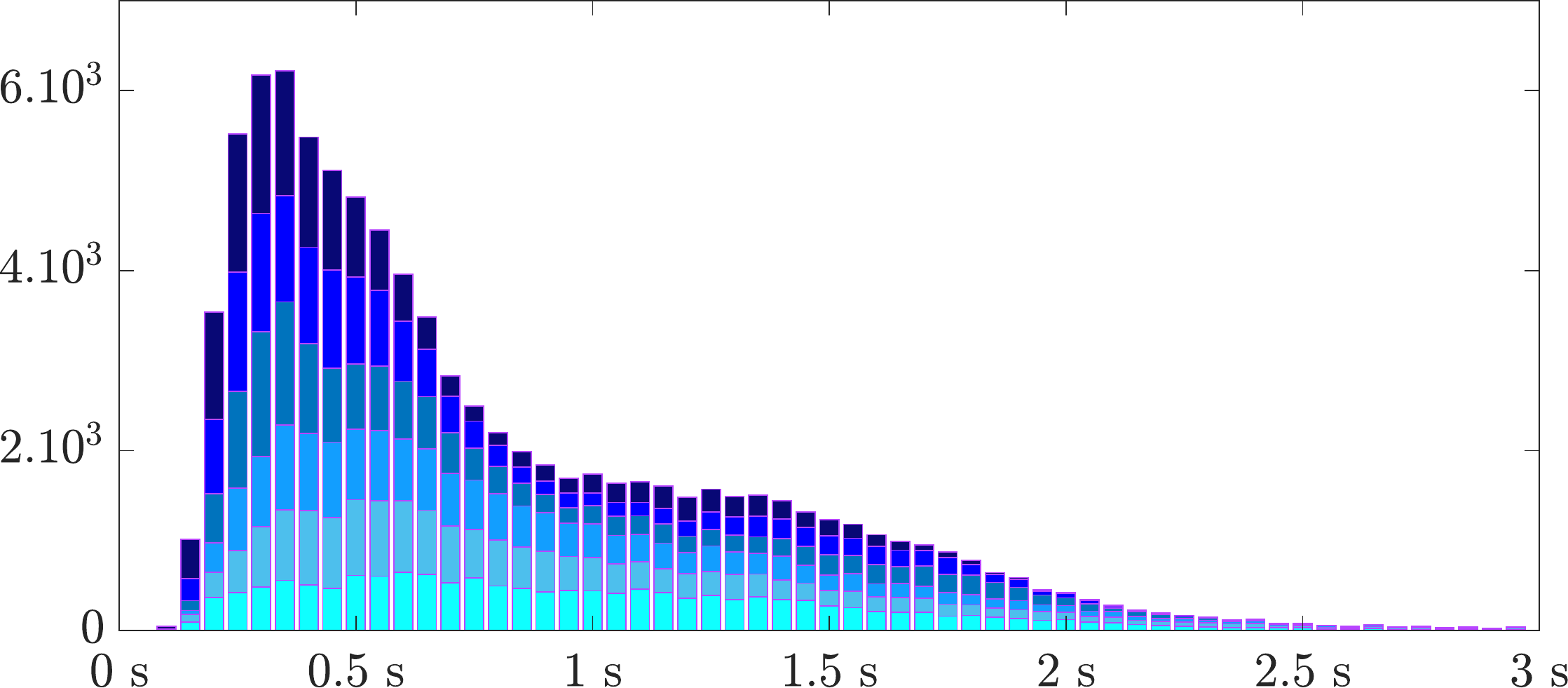}{.49\textwidth}{(d)}\label{fig:hist_RT_RB}\hfill
	}
	\caption{\label{fig:hists} Histograms of $\bar{\alpha}(b)$ [(a),(c)] and $\textrm{RT}_{30}(b)$ [(b),(d)] values in 6 octave bands for 15,000 RIRs using Unif [(a),(b)] vs. RB [(c),(d)] sampling.}
\end{figure*}

Acoustic parameters include the absorption $\alpha_{i}(b)$ and scattering $s_{i}(b)$ coefficients of each of the 6 surfaces $i$ in each of the 6 octave bands $b$. Two different strategies were explored to sample absorption coefficients. The first, most straightforward one, is to draw all 36 coefficients uniformly at random between 0 and 1 for each RIR. We later refer to this approach as \textit{Unif}, which is also the approach employed in the recent paper \cite{yu2020room}. The obtained $\bar{\alpha}(b)$ distribution (Eq.(\ref{eq:alpha})) over 15,000 simulated RIRs is shown in Fig.~\ref{fig:hist_alpha_Unif}. As can be observed in Fig.~\ref{fig:hist_RT_Unif}, the resulting histogram of $\textrm{RT}_{30}(b)$ values\footnote{We denote by $\textrm{RT}_{X}(b)$ a reverberation time calculated on a Schroeder curve's slope from $-5$ to $ -5-X$~dB \cite{Schroeder:65}.} is narrowly spread around 150 ms, which is an unusual value mostly encountered in semi-anechoic chambers. This is because using this technique, drawing four or more reflective absorption profiles within a same room (\textit{e.g.} $\bar{\alpha}_i(b)<0.15$ for all $b$) is very unlikely. Yet, highly reflective profiles are frequently encountered in real buildings. These are characteristics of hard surfaces made of, \textit{e.g.}, concrete, bricks or tiles. The absorption profiles of 26 such materials are plotted in Fig.~\ref{fig:profiles_reflective}. As can be seen, they are all roughly frequency-independent with absorption coefficients below $0.12$. 
Based on this, we designed the following new \textit{Reflectivity Biased} (RB) sampling strategy:
\begin{enumerate}
	\item for each surface type (wall, floor, ceiling), toss a coin;
	\item on heads, draw \textit{reflective} frequency-independent absorption profiles uniformly at random in $[0.01, 0.12]$ for these surfaces;
	\item on tails, draw \textit{non-reflective} frequency-dependent absorption profiles uniformly at random within predefined ranges depending on the surface type (see Fig.~\ref{fig:profiles}).
\end{enumerate}

Note that walls are either all reflective or all non-reflective, but may still have distinct profiles.
The non-reflective ranges are chosen to encompass typical materials used on walls, floors and ceilings in common buildings, as shown in Fig.~\ref{fig:profiles_walls}, \ref{fig:profiles_floors} and \ref{fig:profiles_ceilings}.
As can be seen in Fig.~\ref{fig:hist_RT_RB} and Fig.~\ref{fig:hist_alpha_RB}, the proposed RB sampling technique results in more diverse and more representative distributions for both reverberation  times $\textrm{RT}_{30}(b)$ and mean absorption coefficients $\bar{\alpha}(b)$. \textcolor{black}{The peak around 0.06 observed in Fig.~\ref{fig:hist_alpha_RB} is consistent with the proposed bias towards reflective surfaces and the chosen realistic absorption ranges.}

Finally, for both the Unif and the RB sampling strategies, the same frequency-dependent scattering profile was used for all surfaces. This approach, previously used in \cite{gaultier2017vast}, is based \textcolor{black}{on the interpretation that the diffuse-rain model of Roomsim} globally captures random reflections in the room rather than specific local effects. While random scattering coefficients in $[0,1]$ were used in all octave bands for Unif, we respectively used the ranges $[0,0.3]$ and $[0.2,1]$ for octave bands in \{125~Hz, 250~Hz, 500~Hz\} and \{1~kHz, 2~kHz, 4~kHz\} for RB. This choice is guided by scattering profiles measured in real rooms as reported in \cite{vorlander2000definition}.
Overall, one training set of 15,000 RIRs and one development set of 5,000 RIRs were generated for each of the two sampling techniques.


\section{Neural Network Models and Training}
\label{sec:training} 
\subsection{Data pre-processing}
\label{subsec:pre-proc}
A crucial question in supervised learning is that of finding an appropriate representation for input data, which is sometimes referred to as the \textit{feature extraction} step. Ideally, one seeks a representation that preserves or enhance features that are relevant for estimating the output, while removing unnecessary or redundant ones. In learning-based audio signal processing applications, \textcolor{black}{phase-less} time-frequency representations such as magnitude spectrograms or Mel-Frequency Cepstral Coefficients have been widely used. Since frequency-dependent values are sought, such representations seem attractive at first glance. However, \textcolor{black}{by discarding phase} they would remove fine-grain temporal information such as the timings of early echoes in RIRs. These timings could be exploited to infer geometrical properties of the room that in turn correlate with absorption coefficients conditionally on the reverberation time, as showed by (\ref{eq:sabine}). Alternatively, one could consider \textcolor{black}{invertible} complex time-frequency representations such as the short-term Fourier transform (STFT). Our preliminary experiments in that direction were however not conclusive, possibly due to the difficulty of handling non-linear \textcolor{black}{complex} phase behavior in the networks, or because any choice of STFT parameters implies a non-obvious compromise between time and frequency resolution \textcolor{black}{at each frame.}
Consequently, we choose to let the network learn its own internal representation of time-domain RIRs, in an end-to-end fashion. This approach has recently showed considerable success in other audio signal processing applications, \textit{e.g.}, \cite{luo2018tasnet}.

RIRs obtained by Roomsim were resampled from 48 to 16~kHz. In fact, the highest octave band considered does not exceed 5.7~kHz, suggesting that 12~kHz could be sufficient for our application. However, higher-frequency features such as the times of arrival of early reflections may still carry useful information. On the other hand, overly relying on very high frequencies would be disconnected from real applications, as receivers and emitters used to measure RIRs are always band-limited in practice. Only the first 500~ms of RIRs were preserved, as this range is expected to contain the most salient acoustical information, \textcolor{black}{including both early and late reflections}. This resulted in 8,000-dimensional input vectors. A random white Gaussian noise with signal-to-noise ratio (SNR) 30~dB was also added to every RIR in the datasets. This is expected to make learned models more robust, and to prevent them from relying on vanishingly small values in the RIRs, which would be inaccessible in practical applications. Finally, all input vectors were normalized to have a maximum value of 1. This is done to facilitate learning, and also to prevent models from relying on the RIR's absolute amplitude which is often inaccessible in practical applications due unknown source and microphone gains.

\subsection{Network design}
\label{subsec:architecture}

\begin{figure}[t!]
	\centering
	\figcolumn{
		\fig{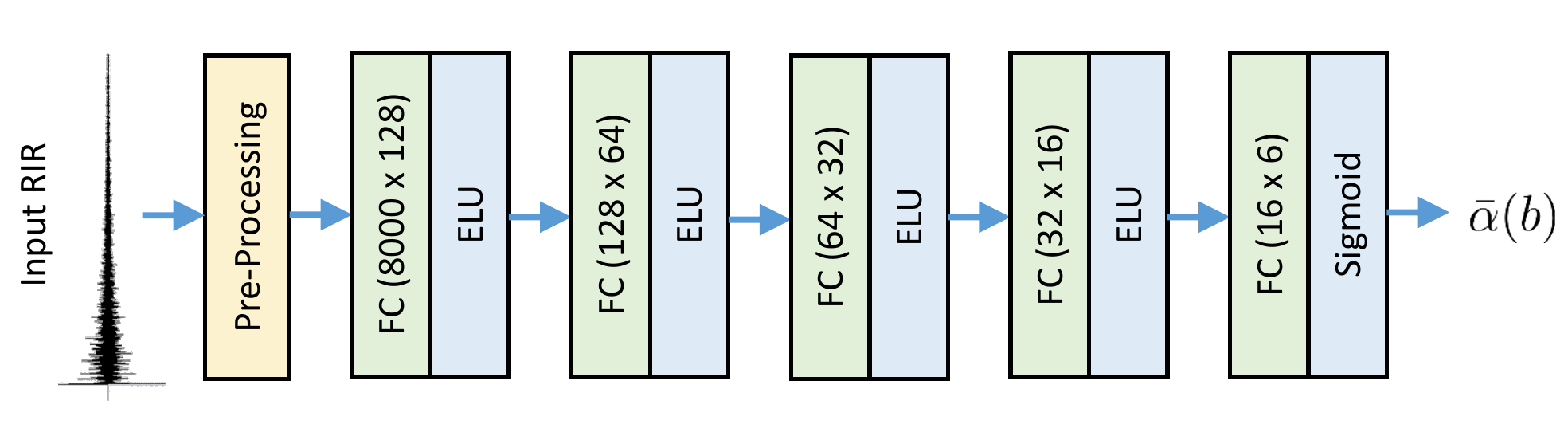}{.50\textwidth}{(a)} \label{fig:MLP}
		\fig{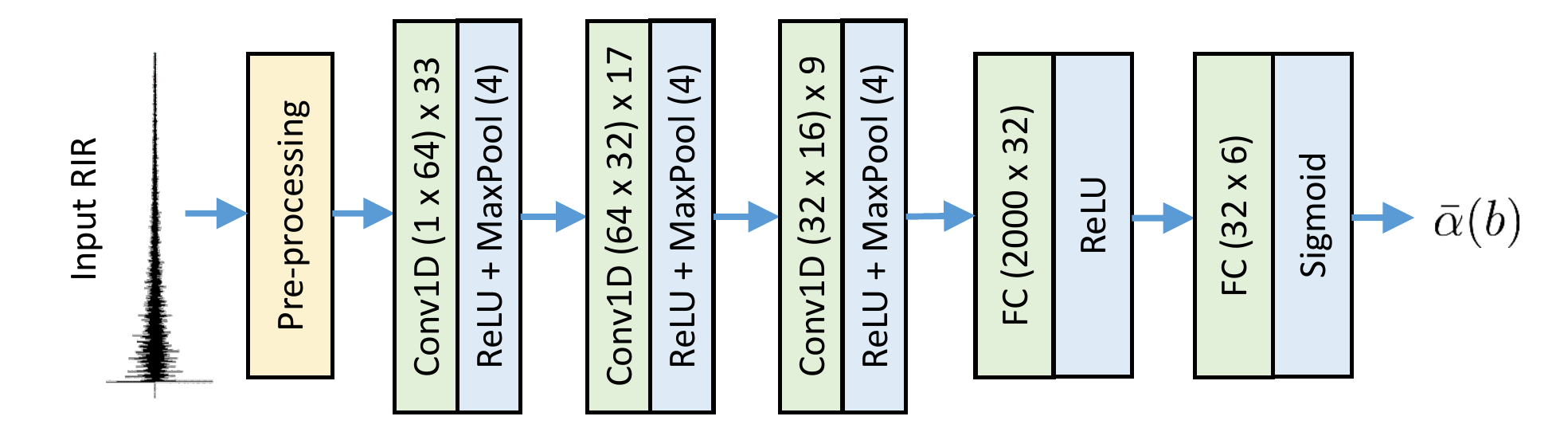}{.50\textwidth}{(b)} \label{fig:CNN}
	}
	\caption{\label{fig:networks} Neural network architectures. (a) Multilayer perceptron (MLP), (b) Convolutional neural network (CNN).}
\end{figure}

Two commonly used neural network architectures are considered for this study, namely, the multilayer perceptron (MLP) depicted in Fig.~\ref{fig:MLP}, and the convolutional neural network (CNN), depicted in Fig.~\ref{fig:CNN}. The MLP is made of three fully connected hidden layers of successive dimensions 128, 64 and 32, each followed by exponential linear units (ELUs). The CNN starts with three consecutive 1D-convolutional hidden layers with a stride of 1, respective filter sizes 33, 17, 9 with \textcolor{black}{zero-padding to preserve dimensionality after each convolution}, and number of filters 64, 32 and 16. Each convolution is followed by a max pooling layer of width 4 and ELUs. The resulting output of dimension 2000 is then passed through a fully connected hidden layer of size 32 with ELUs. \textcolor{black}{This particular designs of layers are meant to define two simple dimensionality-reducing networks of relatively small and comparable size and depth.}
For each network, a final fully-connected output layer is used to yield the desired output vector, evaluated by a mean-squared error loss-function. Networks are optimized on the training set using batches of size of 1000 and ADAM \cite{kingma2014adam} with a learning rate of 0.001. Parameters yielding the lowest average loss on the development set over 400 epochs are used in all experiments. These meta-parameters and choice of ELUs rather than rectified linear units (RELUs) were guided by preliminary experiments on the development sets.

\begin{figure}[!t]
	\centering
	\includegraphics[width=0.32\textwidth]{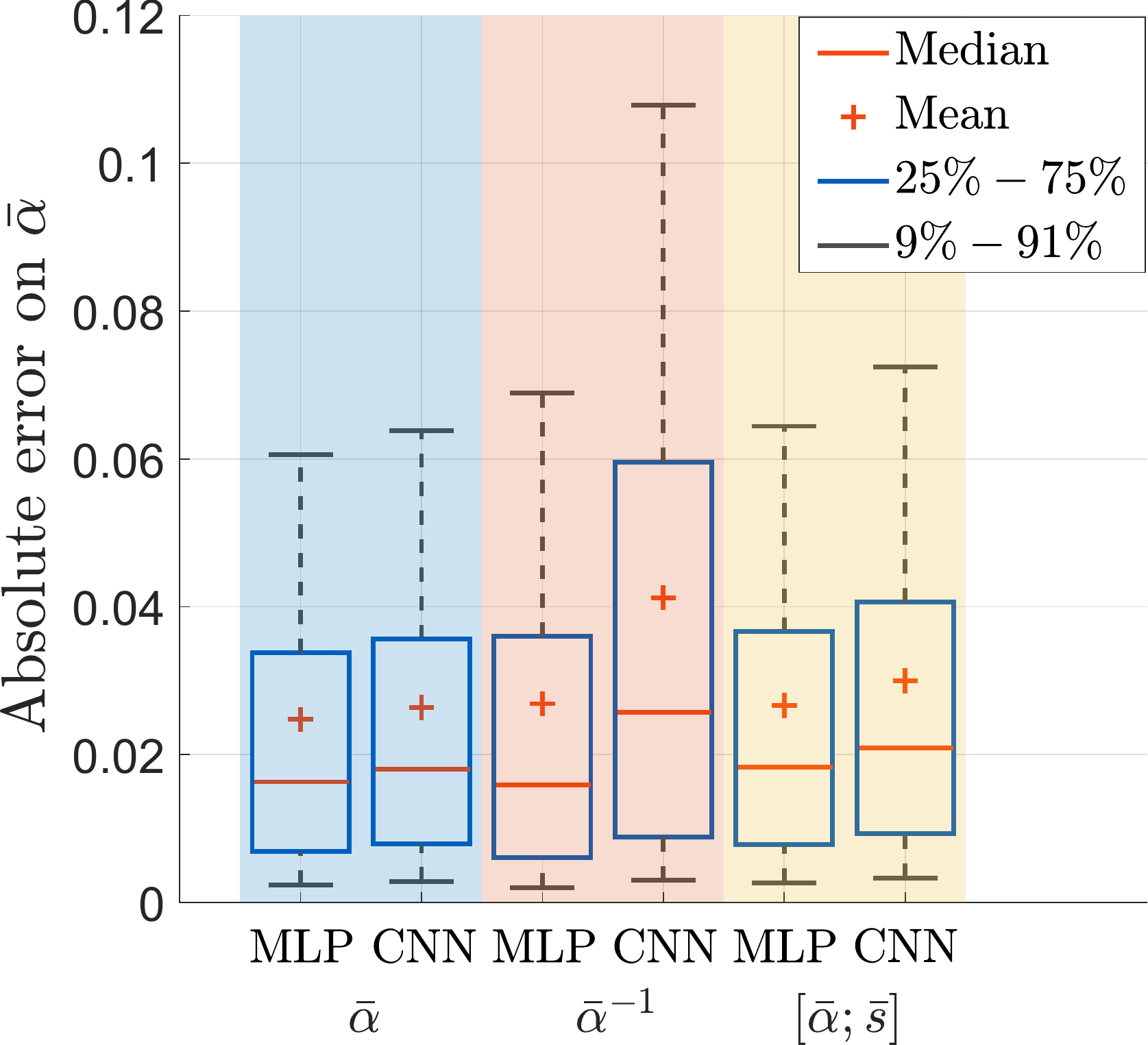}
	\caption{\label{fig:comparison_outputs} Comparison of 3 output layers, trained on RB, evaluated on RB development set.}
\end{figure}

\begin{figure}[!t]
	\centering
	\figline{
		\fig{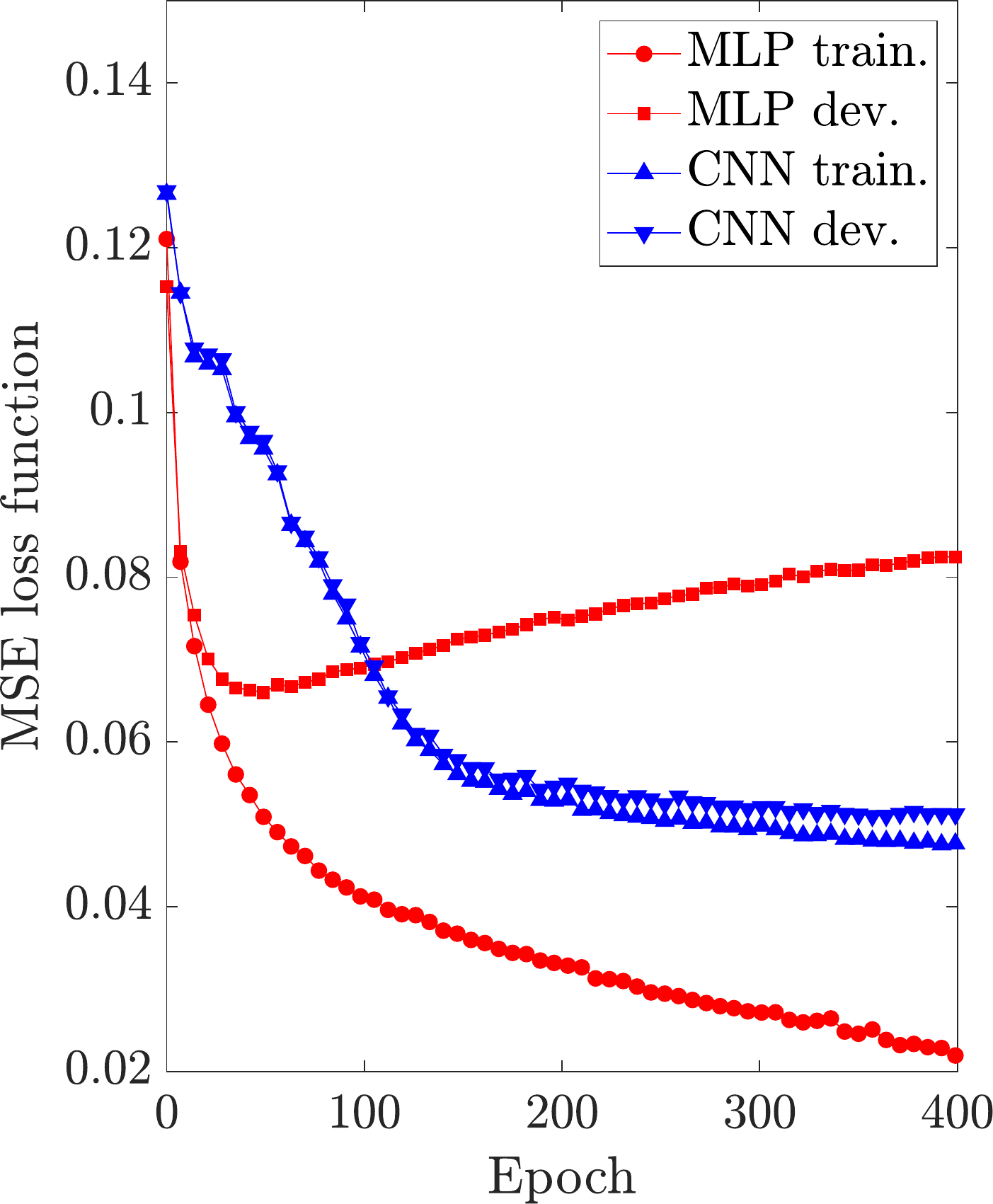}
		{.23\textwidth}
		{(a)}\label{fig:loss_Unif}
		\fig{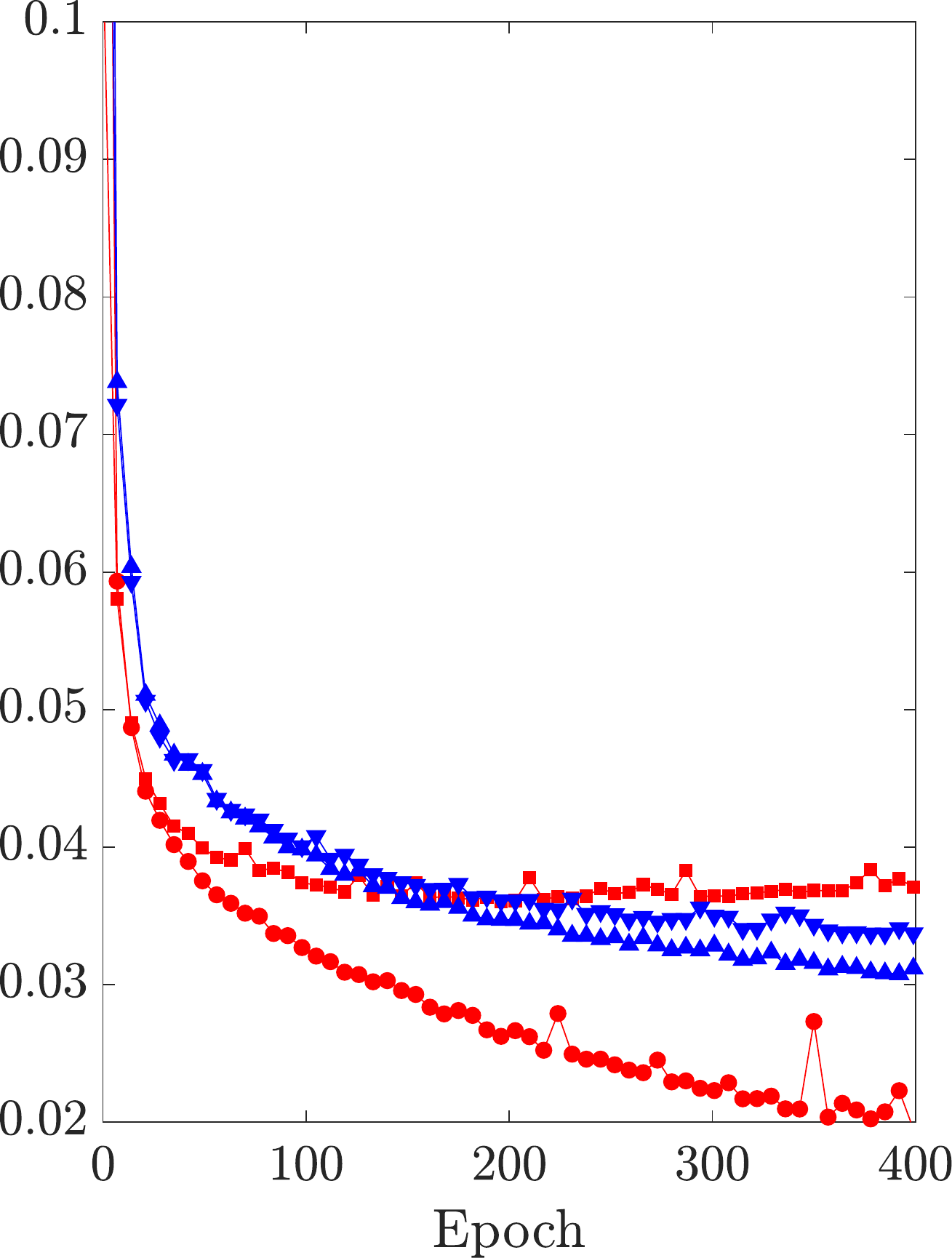}
		{.22\textwidth}
		{(b)}\label{fig:loss_RB}
	}
	\caption{\label{fig:loss}Loss evolution on training and development sets.
		(a) Unif datasets, (b) RB datasets.}
\end{figure}

Three different output targets were considered: (i) the 6-dimensional vector of mean absorption coefficients in all octave bands $\bar{\alphavect} \in [0, 1]^6$, (ii) the vector of inverse mean absorption coefficient $\bar{\alphavect}^{-1}\in\mathbb{R}^{+6}$ or (iii) the concatenation of the mean absorption and scattering coefficients $[\bar{\alphavect};\bar{\boldsymbol{s}}]\in[0, 1]^{12}$.  The second idea derives from the fact that the reverberation of a room is roughly inversely proportional to the mean absorption \textcolor{black}{in DSF conditions, \textit{e.g.}, Sabine's law} \cite{Kuttruff:09}. The third idea is to test whether annotating the network with scattering coefficients at train time could help the estimation of absorption, \textit{i.e.}, multi-task learning. Output values in $[0,1]$ were constrained using sigmoid gates while positive values were constrained using a rectified linear units (ReLU). A comparison of the distribution of absolute errors on $\bar{\alphavect}$ obtained on the development set of RB using these three targets is shown in Fig.~\ref{fig:comparison_outputs}. \textcolor{black}{In the remainder of the article, the absolute error is defined as the absolute difference between target and estimated values. For a given dataset, reported means or box plots are computed over all input RIRs, but also over all 6 octave bands, unless stated otherwise.} As can be seen, using inverse or concatenated vectors yield equivalent or worse results than simply using $\bar{\alphavect}$. Hence, only networks outputting $\bar{\alphavect}$ are considered in the remainder of the paper.

Fig.~\ref{fig:loss_Unif} and \ref{fig:loss_RB} show the evolution of the loss functions of the two networks on the training and development sets for both Unif and RB. It can be observed that the MLP is more prone to over-fitting than the CNN. This suggests that the latter generalizes better to unseen RIRs, an effect which will be confirmed in section \ref{sec:real_results}. This might be explained by the use of temporal convolutions, which may more efficiently capture the global frequency content of RIRs than fully connected layers, while discarding less relevant local information.

\section{\label{sec:results} Experiments and results}

\begin{figure*}[ht!]
	\centering
	\figline{
		\leftfig{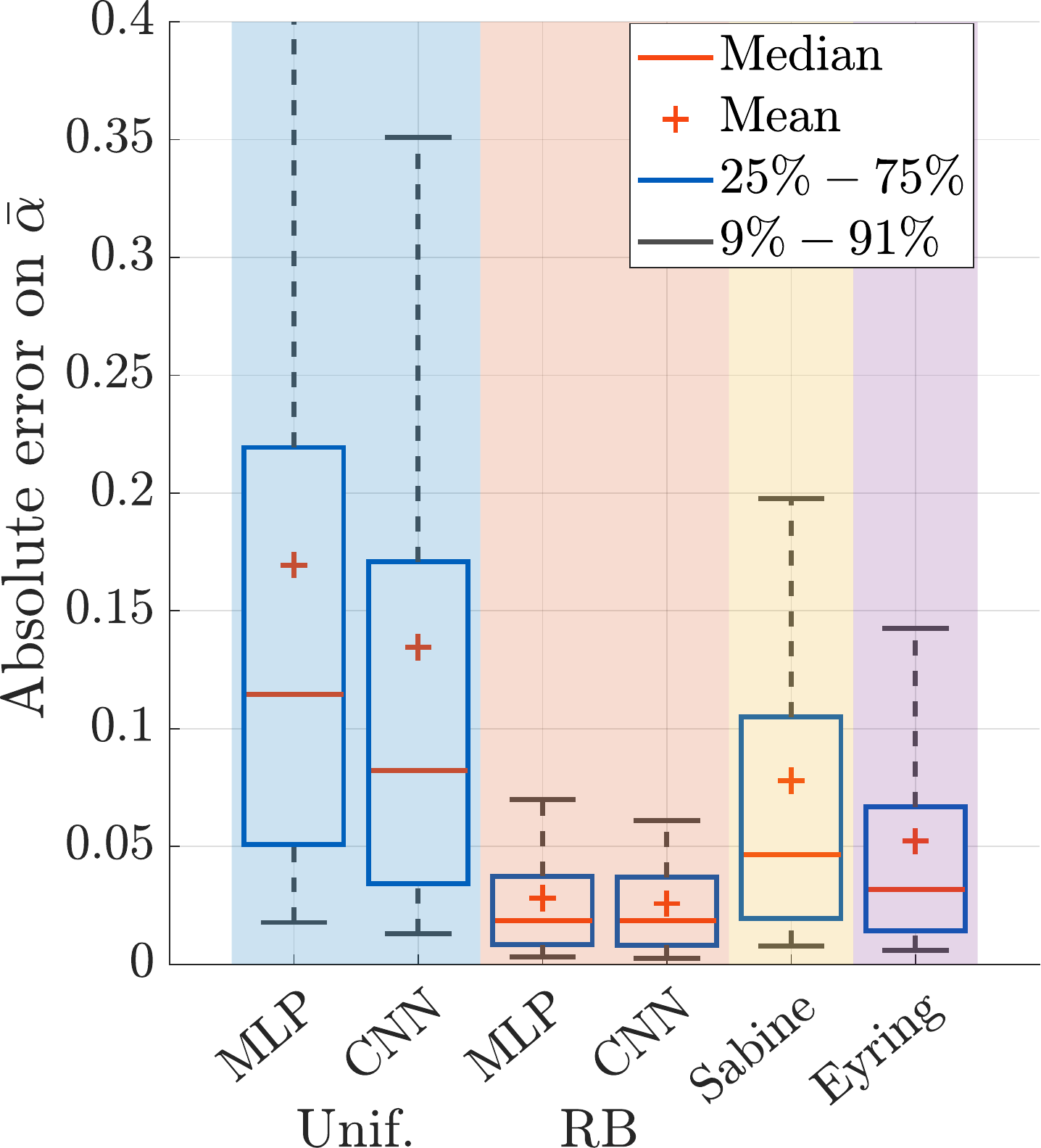}{.31\textwidth}{(a)}\label{fig:err_realistic}
		\leftfig{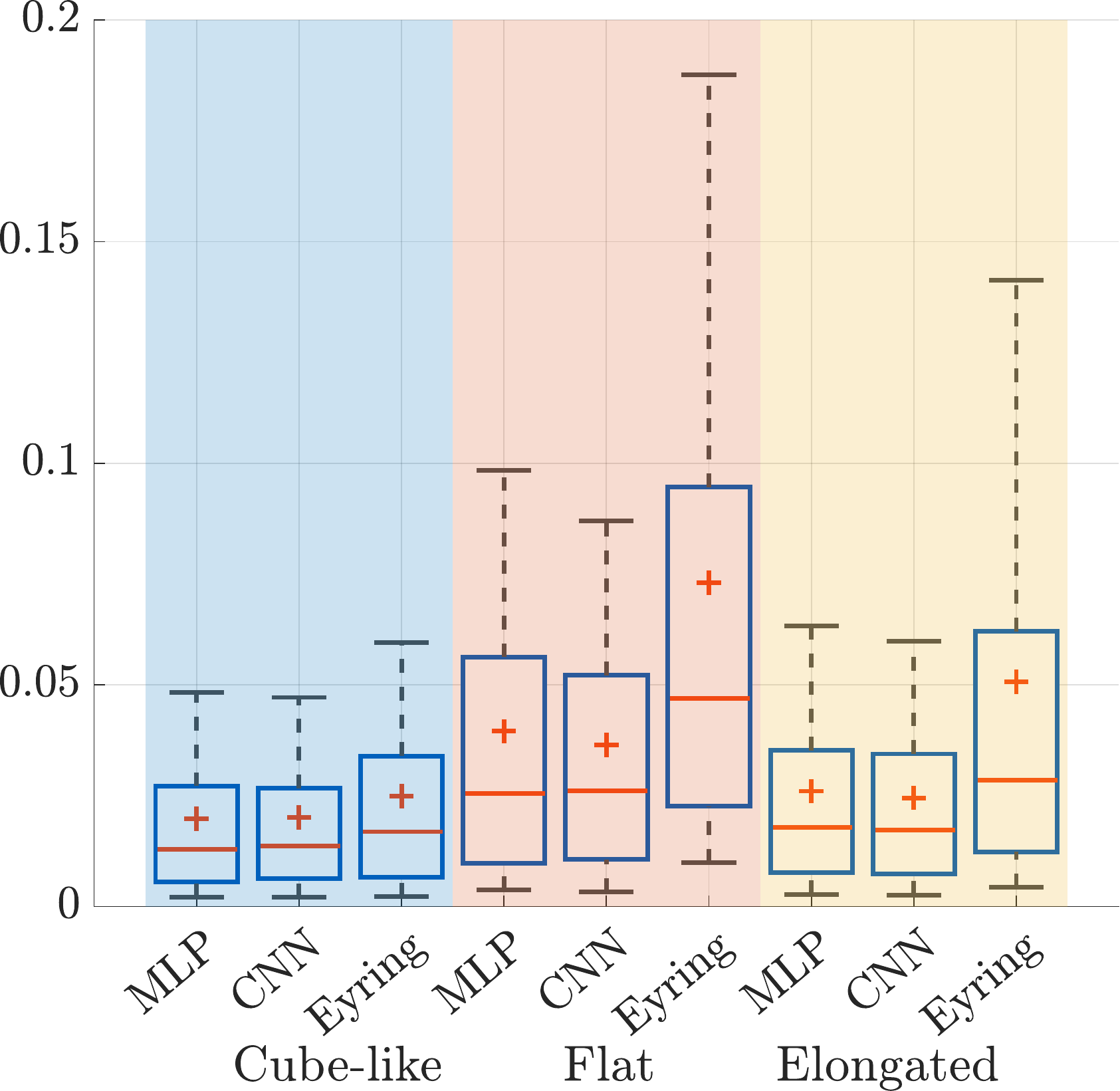}{.35\textwidth}{(b)}\label{fig:err_geo}
		\leftfig{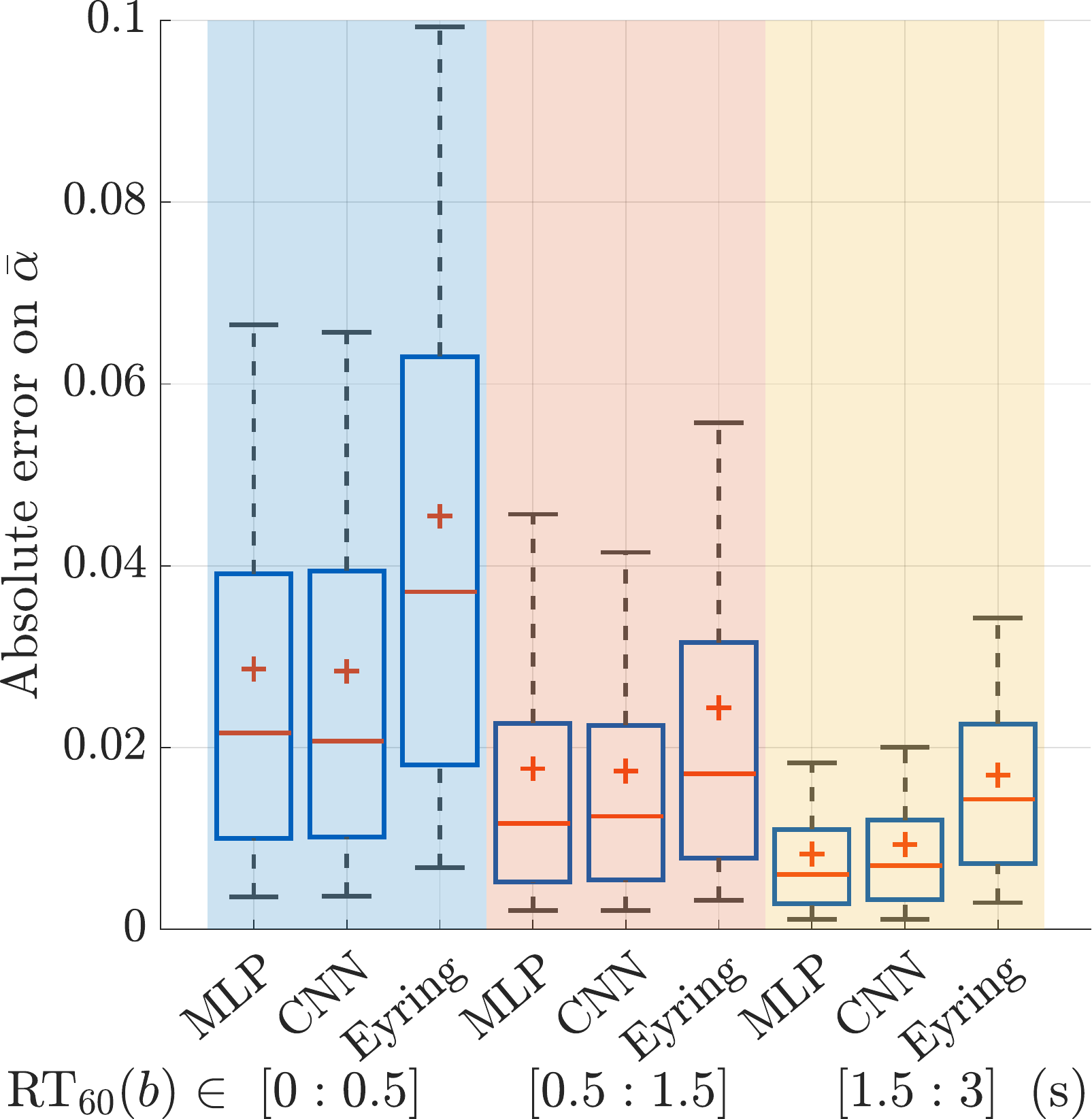}{.33\textwidth}{(c)}\label{fig:err_rev}
		\hfill}
	\figline{
		\leftfig{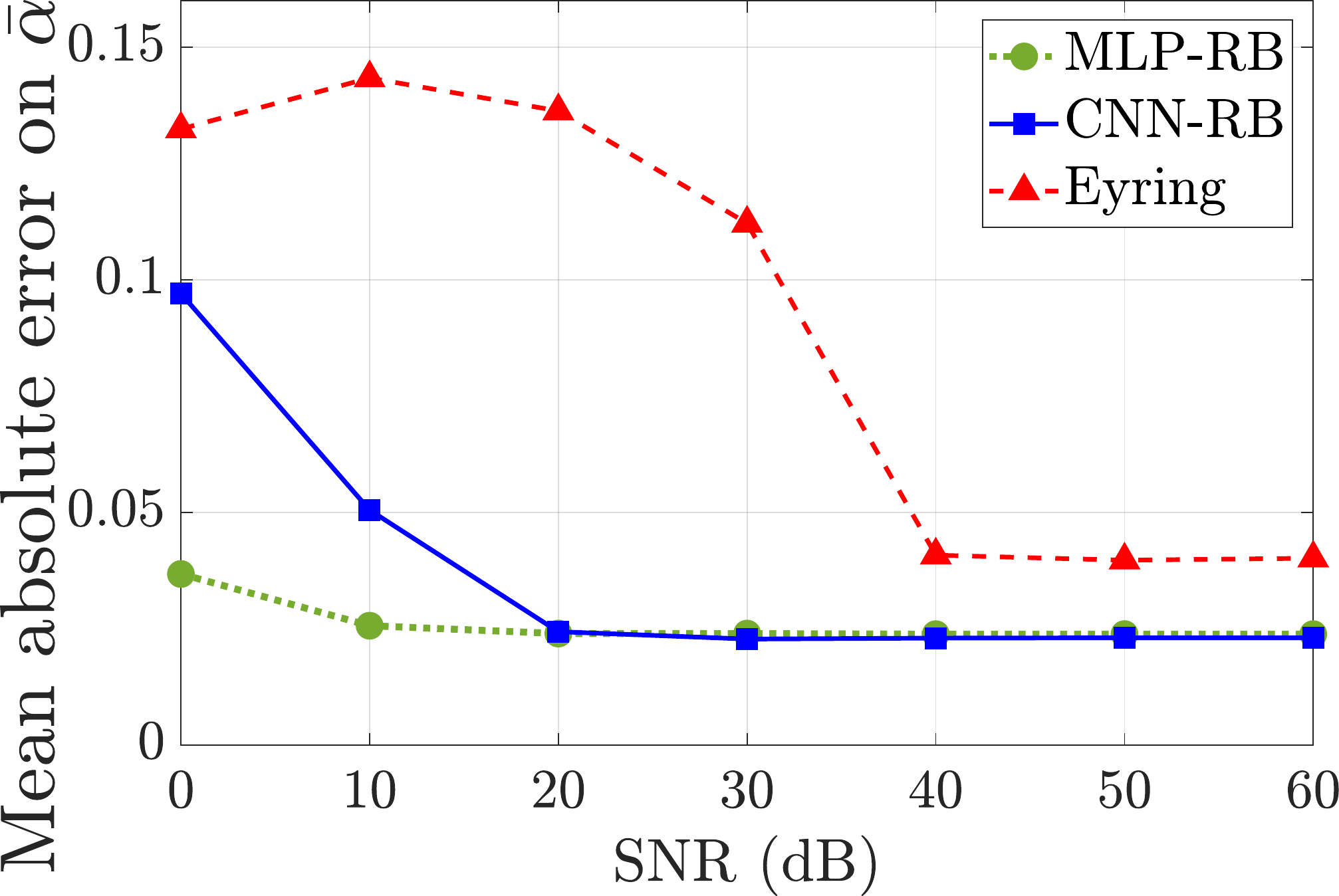}{.33\textwidth}{(d)}\label{fig:err_SNR}
		\leftfig{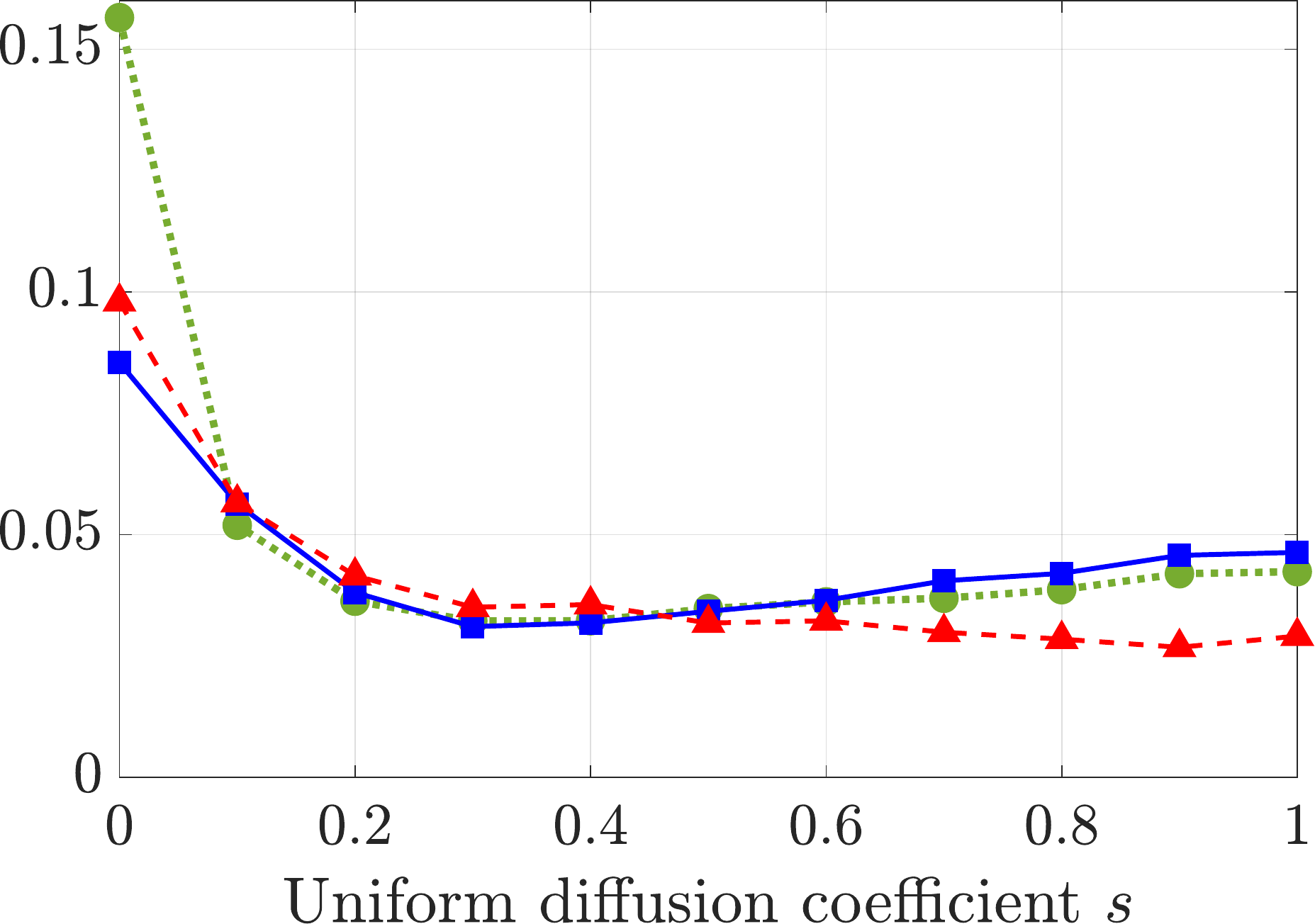}{.31\textwidth}{(e)}\label{fig:err_s}
		\leftfig{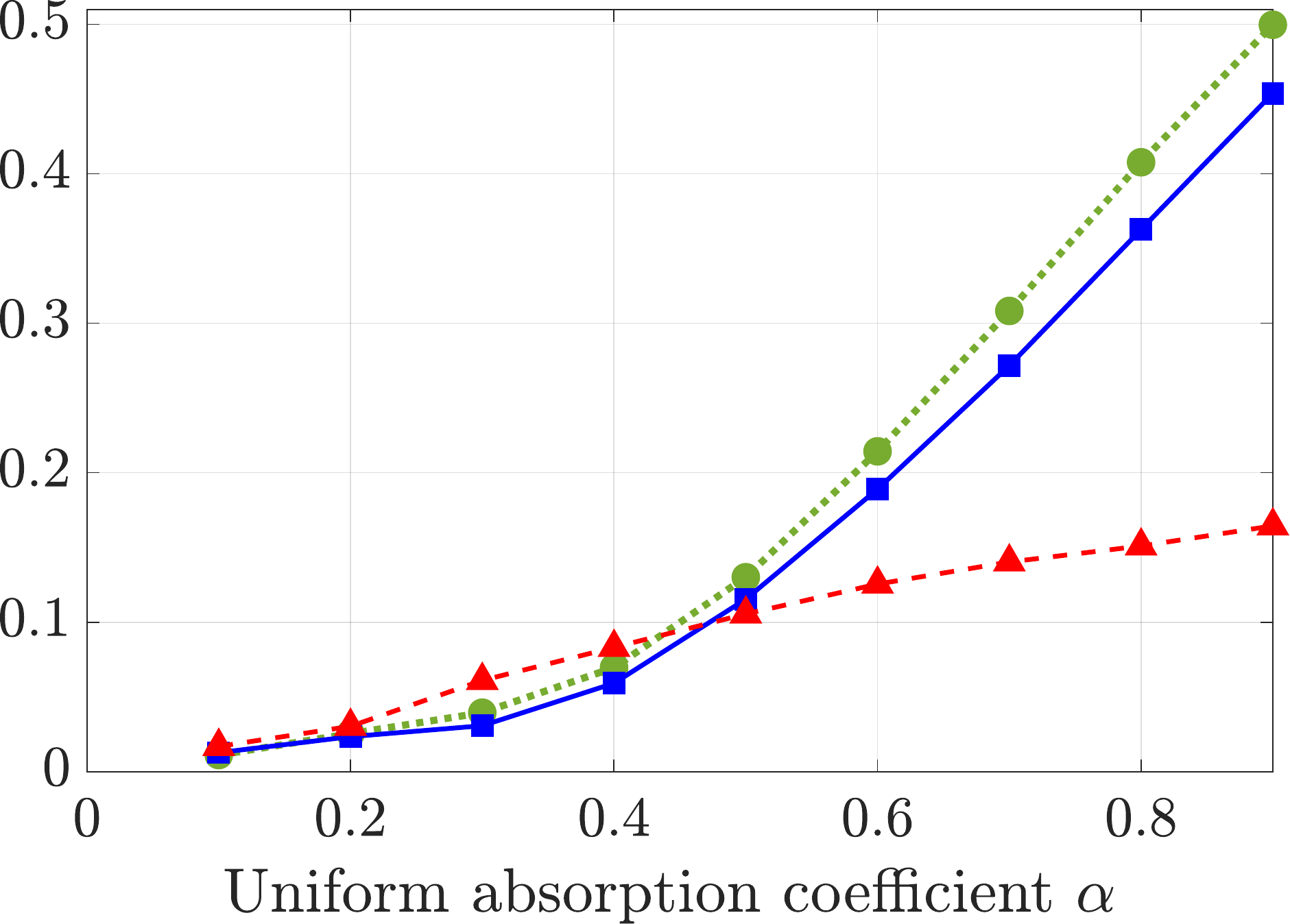}{.31\textwidth}{(f)}\hfill}\label{fig:err_alpha}
	\caption{\label{fig:simulation_results}
		Comparison of $\bar{\alpha}$ estimation errors on different simulated test sets of 500 RIRs each.
		(a) Realistic test set,
		(b) influence of geometry,
		(c) influence of reverberation time,
		(d) influence of noise,
		(e) influence of diffusion,
		(f) influence of absorption.}
\end{figure*}

\subsection{\label{sec:baseline}Baseline models}
\textcolor{black}{As a comparison point with the proposed neural models, we use mean absorption estimates obtained using the well-known Sabine's law and its more precise variant from Eyring from reverberation theory \cite{Kuttruff:09}:
	\begin{align}
		&\bar{\alpha}_{\textrm{Sabine}}(b) = 0.163\cdot V/(S\cdot RT(b)) \label{eq:sabine}\\
		&\bar{\alpha}_{\textrm{Eyring}}(b) = -\ln\left(1-\bar{\alpha}_{\textrm{Sabine}}(b)\right) \label{eq:eyring}
	\end{align}
	where $V$ denotes the room's volume and $S=\sum_i S_i$ its total surface. Eyring's and Sabine's models are always given the true volume $V$ and total surface $S$ of the room in all experiments. Obviously, the diffuse sound field (DSF) hypothesis inherent to these classical models is not theoretically verified for many of the considered room configurations. In order to better understand the impact of this limitation, a preliminary study was carried out on the Unif and RB training databases. The reverberation time used in the formulas was calculated on different dynamics ($[-5~\textrm{dB}, -15~\textrm{dB}]$, $[-5~\textrm{dB},-20~\textrm{dB}]$, $[-5~\textrm{dB}, -25~\textrm{dB}]$, $[-5~\textrm{dB}, -35~\textrm{dB}]$, $[-5~\textrm{dB}, -65~\textrm{dB}]$) of the Schroeder curves \cite{Schroeder:65} and the resulting distributions of absolute errors were estimated. The dynamic $[-5~\textrm{dB}, -35~\textrm{dB}]$, \textit{i.e.}, $\textrm{RT}_{30}(b)$, was retained for our study as it offered the smallest median values of absolute errors for the Unif and RB training databases, \textit{i.e.}, 0.07 and 0.03 respectively. Such low errors show that the exploitation of these DSF-based models in our comparative study, while limited, is not unreasonable for the selected room configurations.}


\subsection{Simulation results} 
We now compare the different learned models (MLP-Unif, MLP-RB, CNN-Unif, CNN-RB) to Eyring's (\ref{eq:eyring}) and Sabine's (\ref{eq:sabine}) models on the task of estimating surface-weighted mean absorption coefficients (\ref{eq:alpha}) from a simulated RIR. A variety of simulated test sets, containing 500 RIRs each and all generated with Roomsim, are considered.

\note{\sout{and the metric used is the mean absolute difference between estimated mean absorption coefficients and true ones, defined by (\ref{eq:alpha}). Eyring's and Sabine's formulas are always given the true volume $V$ and total surface $S$ for each room. Moreover, a preliminary study was made on a noiseless version of the RB training set to identify the best $\mathrm{RT}(b)$ computation method for these techniques, namely, what log-energy decay $\Delta E$ should be considered to compute the slope of Schroeder's curves, starting -5~dB from the initial level. We tested the values $\Delta E = -10, -15, -20, -30$ and $-60$~dB, and selected $\Delta E=-30$~dB, which yielded the smallest mean estimation errors.}}

\begin{figure*}[!t]
	\centering
	\begin{minipage}{0.68\textwidth}
		\centering
		\includegraphics[height=0.26\textheight]{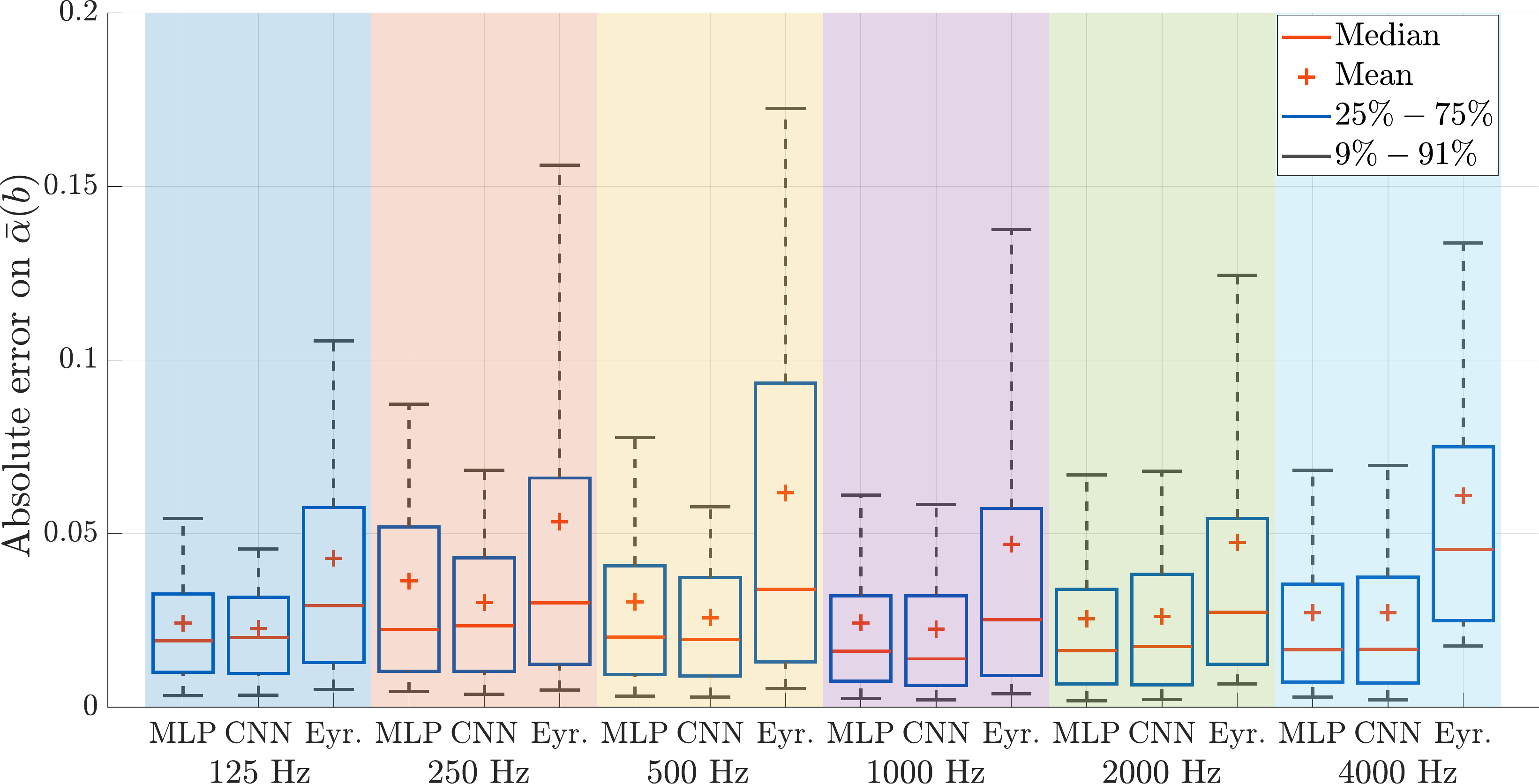}
		\caption{\label{fig:err_octave}Comparison of $\bar{\alpha}(b)$ estimation errors on the realistic test set in different octave bands. The set used for training networks is RB.}
	\end{minipage}
	\hfill
	\begin{minipage}{0.29\textwidth}
		\centering
		\includegraphics[height=0.26\textheight]{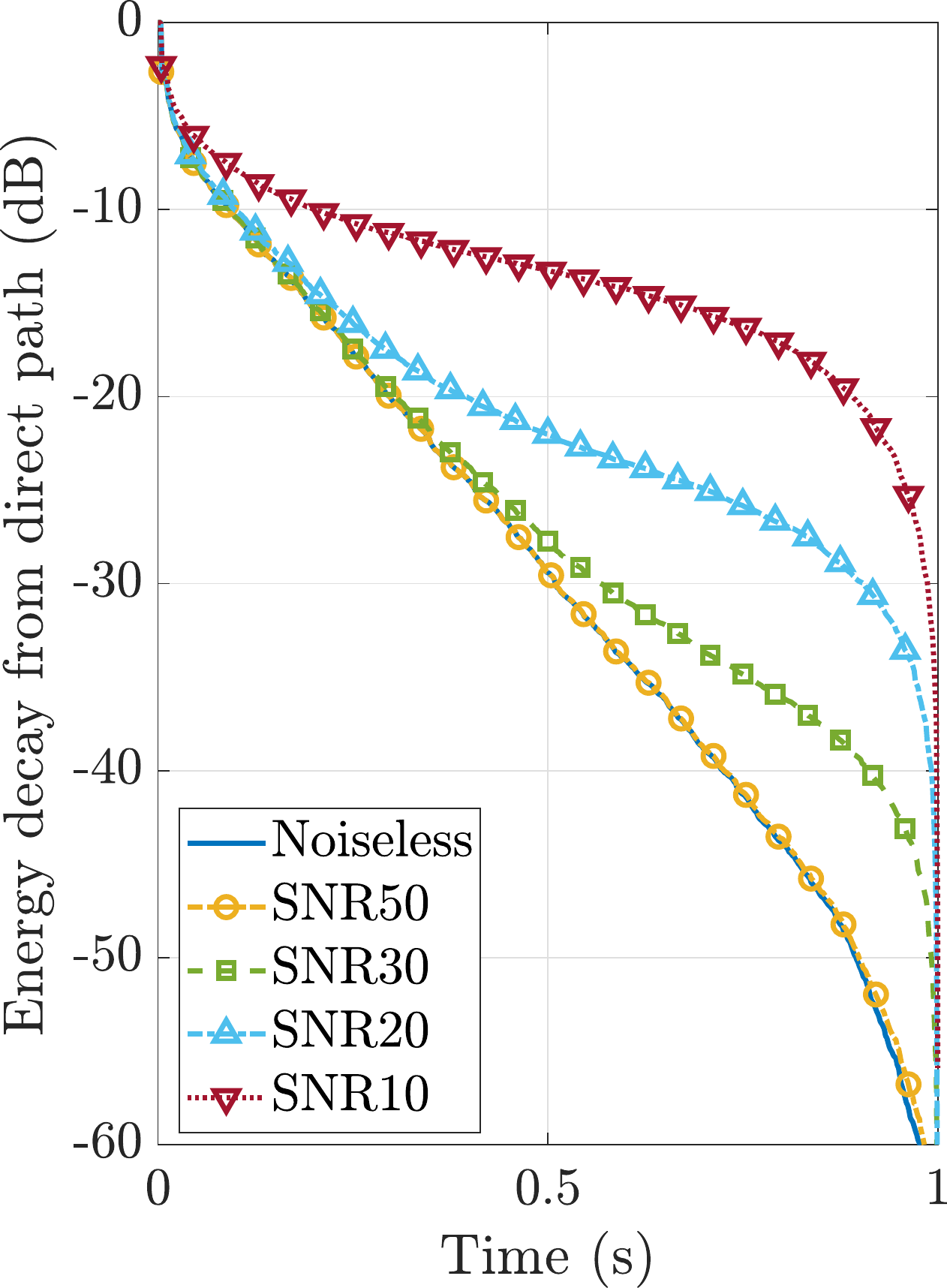}
		\caption{\label{fig:schroeder_curves}1 kHz Schroeder curves of a RIR under varying SNRs.}
	\end{minipage}
\end{figure*}

The first simulated test set, called \textit{realistic}, only contains surface materials commonly encountered in real buildings, drawn uniformly at random from the database presented in Fig.~\ref{fig:profiles}. Five fixed geometries representative of typical rooms were selected for this set with the following $(L_x, L_y, L_z)$ dimensions in meters: $(4, 5, 3)$, $(10, 2, 3)$, $(10, 5, 3)$, $(5, 8, 2.5)$, $(10, 10, 5)$. The scattering of the walls and the noise level is the same as in RB datasets. Absolute errors obtained with the 6 methods are presented in the form of box plots in Fig.~\ref{fig:err_realistic}. As can be seen, networks trained on the naive Unif training set do not succeed in outperforming classical approaches based on reverberation theory. However, mean estimation errors twice smaller than Eyring's method and with much less variance are obtained using the networks trained on the RB set. As expected, Sabine's estimates show to be slightly less accurate than Eyring's. Hence, results from Unif-trained networks and from the Sabine's model will no longer be reported in what follows. The absolute error distribution was also observed per octave band for this test set (Fig.~\ref{fig:err_octave}). No major differences in errors were observed across octave bands for the different methods. Hence errors will systematically be aggregated over all octave bands in the remainder of this section.

We then conduct a series of experiments on specially crafted simulated test sets to further test the efficiency of the different models against various acoustical conditions. \textcolor{black}{Unless stated otherwise, acoustic parameters follow RB sampling (see Section \ref{subsec:represent}) and RIRs have undergone the same pre-treatment as Section \ref{subsec:pre-proc}.}  
First, Fig.~\ref{fig:err_geo} compares results on three test sets respectively containing only cube-like rooms ($L_x, L_y \in [2,4];\,L_z=2.5$), flat rooms ($L_x, L_y \in[8,10];\,L_z=2.5$) and elongated rooms ($L_x \in[2,4];\,L_y \in[8,10];\,L_z=2.5$).
\note{\sout{While learned models only provide minor improvements over Eyring's under cube-like geometries where the diffuse sound-field assumption is mostly met, they offer a clear advantage in non-homogeneous conditions.}}
\textcolor{black}{Unsurprisingly, with Eyring's model, the smallest absolute errors are obtained on cube-like rooms for which the sound field is closest to diffuse \cite{Hodgson94,Hodgson96}. Logically, both the mean and variance of this error increases for the two other geometrical configurations. While learned models only provide minor improvements over Eyring's formula under cube-like geometries where the DSF assumption is mostly met, they offer a clear advantage in non-homogeneous conditions.}

\note{\sout{Next, Fig.~\ref{fig:err_rev} shows estimation errors on three subsets of a RB-sampled dataset, containing RIRs with constrained reverberation times across all octave bands. It shows that errors decrease when the reverberation time increases. This is expected, as more signal is then available to estimate energy decays. Nevertheless, learned models remain superior to Eyring's under all reverberation conditions.}} 

\textcolor{black}{Fig.~\ref{fig:err_rev} compares the results for three test sets, each associated with a specific reverberation (slightly reverberant, semi-reverberant, reverberant). While obtained errors tend to increase as the reverberation time decreases, learned models remain superior to Eyring's in all conditions. For Eyring, this increase is expected as more reverberant rooms are closer to the DSF hypothesis \cite{Hodgson94,Hodgson96}.}

Fig.~\ref{fig:err_SNR} reports errors as a function of SNR, when additive white Gaussian noise is added to RIR signals (SNR levels are calculated on the first 500ms of RIRs). It can be seen that the Eyring model's estimations degrade abruptly for SNRs of 30~dB or lower. To investigate this effect, Fig.~\ref{fig:schroeder_curves} shows the 1~kHz Schroeder curves of an example RIR under varying noise levels. As can be seen, as the noise level increases, a clean, linear, -30~dB log-energy decay may no longer be available, thus degrading the $\textrm{RT}_{30}$ estimation. This is a well known limitation of reverberation-based techniques, which often require to manually adapt the decay level used depending on measurements. On the other hand, the learned MLP-RB and CNN-RB models, trained on a noisy dataset (30~dB SNR), prove to be much more robust to noise, \textcolor{black}{suggesting that they adaptively extract relevant cues from RIRs}.

Finally, Fig.~\ref{fig:err_s} and \ref{fig:err_alpha} report errors as a function of $\bar{\alpha}$ and mean scattering coefficient $\bar{s}$, where each coefficient is fixed to a constant value across all octave bands and surfaces in each test set. \textcolor{black}{Once again, the behaviour of Eyring's model matches the one expected from reverberation theory, since rooms containing high-scattering, low-absorption materials tend to feature more diffuse sound fields \cite{Hodgson94,Hodgson96}.} On the other hand, learned models perform similarly or better than Eyring's model for $\bar{s}<0.5$ and $\bar{\alpha}<0.5$, but significantly less well otherwise. This is because mean scattering values outside those ranges were not present in the RB training set (see Fig.~\ref{fig:hist_alpha_RB}). While learning-based methods show remarkable interpolation capabilities, they are known to have limited extrapolation capabilities.

\textcolor{black}{To get further insight on the influence of scattering coefficients and diffusion when training neural networks, we tried retraining the CNN model on a purely specular RB set, \textit{i.e.}, using only the image-source method in Roomsim while disabling the diffuse-rain algorithm, as done in, \textit{e.g.}, the learning-based absorption estimation technique  proposed in \cite{yu2020room}. The obtained mean absolute error on $\bar{\alphavect}$ on the realistic test set was 0.18, which is six times larger than when using the original RB set with diffusion activated (0.03). This strongly highlights the importance of taking into account scattering effects when training learning-based acoustic estimation techniques}.



Overall, this extensive simulated study reveals that carefully-trained virtually-supervised models can consistently and significantly outperform conventional reverberation-based techniques \textcolor{black}{in the task of estimating the quantity $\bar{\alphavect}$}, particularly under noisy or non-diffuse sound field conditions. \textcolor{black}{This was expected as the use of Eyring's model is theoretically inadequate under such conditions, even if observed absolute errors were reasonable in practice (see Section \ref{sec:baseline}). In conditions close to the DSF hypothesis, learned models and reverberaton-based models become comparable. This suggests that trained models learned a correction with respect to classical models under non-DSF conditions, by extracting richer features from the RIRs than the mere reverberation times.}

\section{Test on real data}
\label{sec:real_results} 

\subsection{Real dataset}
To evaluate the generalizability of the proposed approach to real measured RIR, we us a subset of the dEchorate dataset~\cite[paper under review]{di2021dechorate}. The dataset consists of RIR measured in a $6~\textrm{m} \times 6~\textrm{m} \times 2.4~\textrm{m}$ acoustic room in the Acoustic Lab of the Bar-Ilan University. The wall and ceiling absorption properties can be changed by flipping double-sided panels with one reflective and one absorbing face.
\note{\sout{This allows obtaining different room configurations, each characterized by a different prominence of early reflections and reverberation levels.}}

Ten different room configurations are considered. They are represented as binary strings of 6 bits in Table~\ref{tab:rooms}, where 1 denotes a reflective surface, 0 an absorbing surface, and the ordered bits respectively represent the floor, the ceiling and the West, South, East and North walls. For each configuration, 90 RIRs from all combinations of 3 sources and 30 receivers spread inside the room are measured. The sources are Avantone Pro Active Mixcube loudspeakers \textcolor{black}{(directional)} and the receivers are AKG CK32 omnidirectional microphones. While room configurations 1 to 9 only contain the sources and receivers, room configuration 10 also contains some typical meeting room furnitures, namely, a table, some chairs and a coat hanger. Each RIR is measured using the exponential sine sweep technique described in~\cite{farina2007advancements}. In this experiment, the octave bands centered at 125~Hz and 250~Hz will not be considered, because the measured RIRs did not exhibit sufficient power in those bands for reliable $\textrm{RT}(b)$ estimations. This observation is consistent with the frequency response provided by the loudspeakers' manufacturer, which decays exponentially from 200~Hz downwards.

\begin{figure*}[!t]
	\centering
	\includegraphics[width=0.8\textwidth]{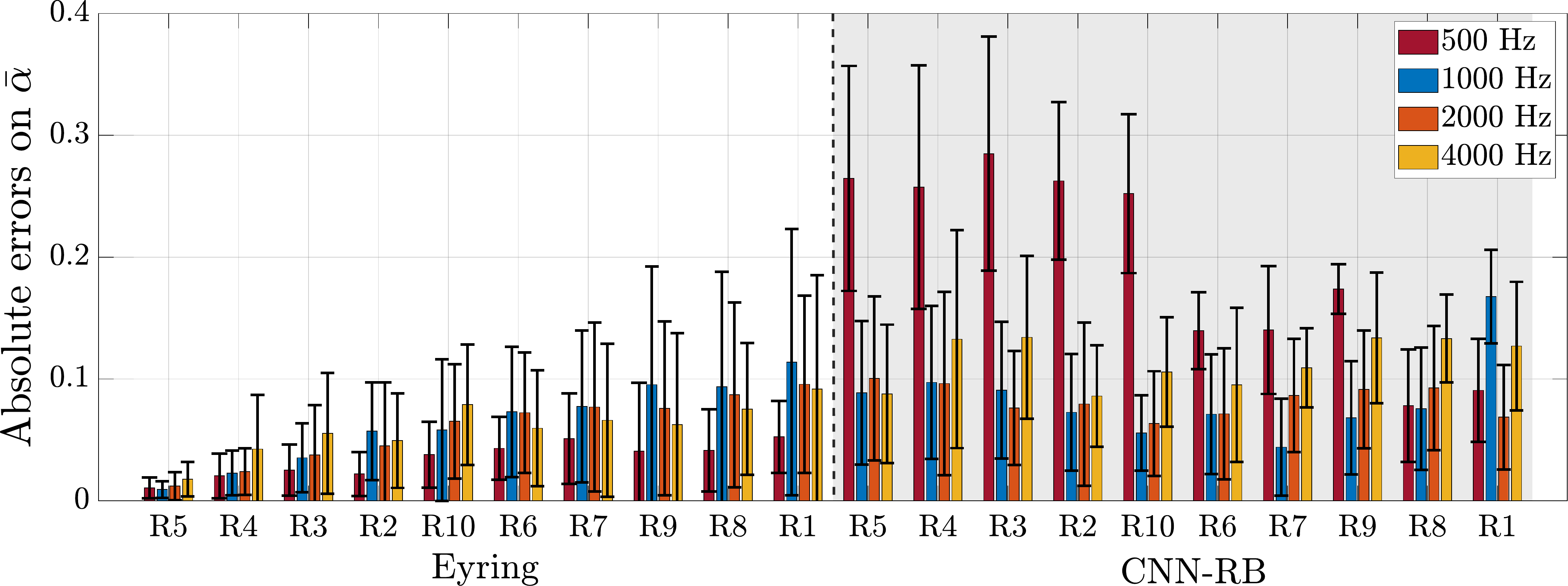}
	\caption{Comparison of $\bar{\alpha}(b)$ mean estimation errors over measured RIRs in 10 rooms and 4 octave bands with Eyring and CNN-RB. Only selected RIRs with Schroeder curves in $\mathcal{A}$ are included.}
	\label{fig:real_results_all}
\end{figure*}

\subsection{Reference absorption values}
A major difficulty in evaluating the considered models on real \textit{in situ} measures is the unavailability of ground truth for the mean absorption coefficients, which would require to know the true absorption profile of every material in the room. While some of them could be inferred from manufacturer's data, only coarse values of $\bar{\alpha}(b)$ would be obtained in this way. 
\textcolor{black}{To overcome this difficulty while ensuring that a single, stable and reliable mean absorption profile is used as a reference for each room, we propose a technique based on the aggregation of multiple RIR measurements.}

For each room configuration, the Schroeder curves of the 90 measured RIRs in 4 octave bands were traced \cite{Schroeder:65}. Then, the Schroeder curves were visually inspected and separated into two sets. Set $\mathcal{A}$ contains Schroeder curves featuring a sufficient linear log-energy decay from -5~dB to -15~dB at least. Set $\mathcal{B}$ contains all the other curves. In practice, $49\%$ of the 3600 Schroeder curves were discarded to the set $\mathcal{B}$ in this way. These mostly corresponded to challenging measurement situations contained in the dEchorate dataset, such as a receiver near a surface, or a loudspeaker facing towards a surface and away from receivers.
Then, for each room configuration and each octave band $b$, the reference mean absorption coefficient $\bar{\alpha}_{\textrm{ref}}(b)$ is taken to be the median value of Eyring's model based on the $\textrm{RT}_{10}(b)$ computed from Schroeder curves in $\mathcal{A}$ only, and on the known room's volume and total surface. This median value $\bar{\alpha}_{\textrm{ref}}(b)$ is taken over at least 5 and on average 47 estimates (see Table~\ref{tab:rooms}), yielding a reliable and robust value. As can be seen in Table~\ref{tab:rooms}, a diversity of mean absorption coefficients $\bar{\alpha}_{\textrm{ref}}(b)$ between 0.12 and 0.52 is represented. This matches quite well the range of values considered in this study (see Fig.~\ref{fig:profiles} and \ref{fig:hist_alpha_RB}).

\note{\sout{Note that aggregating measures from multiple source-receiver pairs in a room is a commonly used technique in room acoustics. A summary of ground truth mean absorption coefficients obtained in this way in the 4 considered octave bands for each of the 10 rooms is provided in Table~\ref{tab:rooms}.}}

\textcolor{black}{To further validate this choice of reference value, the left part of Fig.~\ref{fig:real_results_all} shows the means and standard deviations (stds) of absolute differences between single-RIR Eyring estimates and the proposed median-based reference for each room configuration and each octave band, using RIRs from set $\mathcal{A}$ only. Rooms are sorted left-to-right from the most reverberant one to the least reverberant one. It clearly appears that both the means and stds of differences between single and median-based estimates increase as the reverberation time decreases, consistently with reverberation theory \cite{Hodgson94,Hodgson96}. Nevertheless, both these means and stds remain reasonably low (below 0.1) under all configurations, despite measurements being taken from many different source-receiver placements in the room. This validates our premise of a close-to-diffuse sound field in these experiments, at least when restricting to RIRs inside of the set $\mathcal{A}$ for each octave band.} 

\begin{table*}[t!]
	\begin{center}
		\vspace{-5mm}
		\caption{\label{tab:rooms} Absorption coefficients $\bar{\alpha}_{\textrm{ref}}(b)$ calculated in the 10 room configurations. For each coefficient, the number of corresponding Schroeder curves in $\mathcal{A}$ used to compute the median Eyring's estimate is given in parentheses. Room 10 contains furniture.}
		\begin{ruledtabular}
			\footnotesize
			\begin{tabular}{l|cccccccccc}
				& Room 1 & Room 2 & Room 3 & Room 4 & Room 5 & Room 6 & Room 7 & Room 8 & Room 9 & Room 10 \\
				Config. & 000000 & 011000 & 011100 & 011110 & 011111 & 001000 & 000100 & 000010 & 000001 & 010001\\
				500 Hz  & 0.42 (11) & 0.23 (7) & 0.20 (20) & 0.17 (51) & 0.13 (48) & 0.39 (8) &	0.38 (5) &	0.40 (8) & 0.35 (7) & 0.23 (12) \\
				1000 Hz & 0.52 (62) & 0.28 (83) & 0.25 (86) & 0.17 (89) & 0.13 (90) & 0.44 (79) & 0.41 (74) & 0.44 (69) & 0.43 (70) & 0.33 (72) \\
				2000 Hz & 0.50 (65) & 0.34 (81) & 0.30 (86) & 0.19 (82) & 0.14 (88) & 0.44 (74) & 0.42 (64) &	0.44 (66) & 0.44 (67) & 0.37 (69) \\
				4000 Hz & 0.37 (15) & 0.35 (17) & 0.29 (22) & 0.16 (16) & 0.12 (29) & 0.38 (17) & 0.33 (12) & 0.32 (14) & 0.34 (18) & 0.32 (14) 
			\end{tabular}
		\end{ruledtabular}
	\end{center}
\end{table*}

\subsection{Real data results}
On real RIRs, the MLP models appeared to perform significantly less well than CNN models, yielding errors up to twice larger. This is consistent with the better generalization capabilities of the CNN models observed in Fig.~\ref{fig:loss} and discussed in section \ref{subsec:architecture}. We hence omit the MLP results in the remainder of this section, for compactness.

\begin{figure*}[!t]
	\begin{minipage}{0.54\textwidth}
		\includegraphics[height=0.25\textheight]{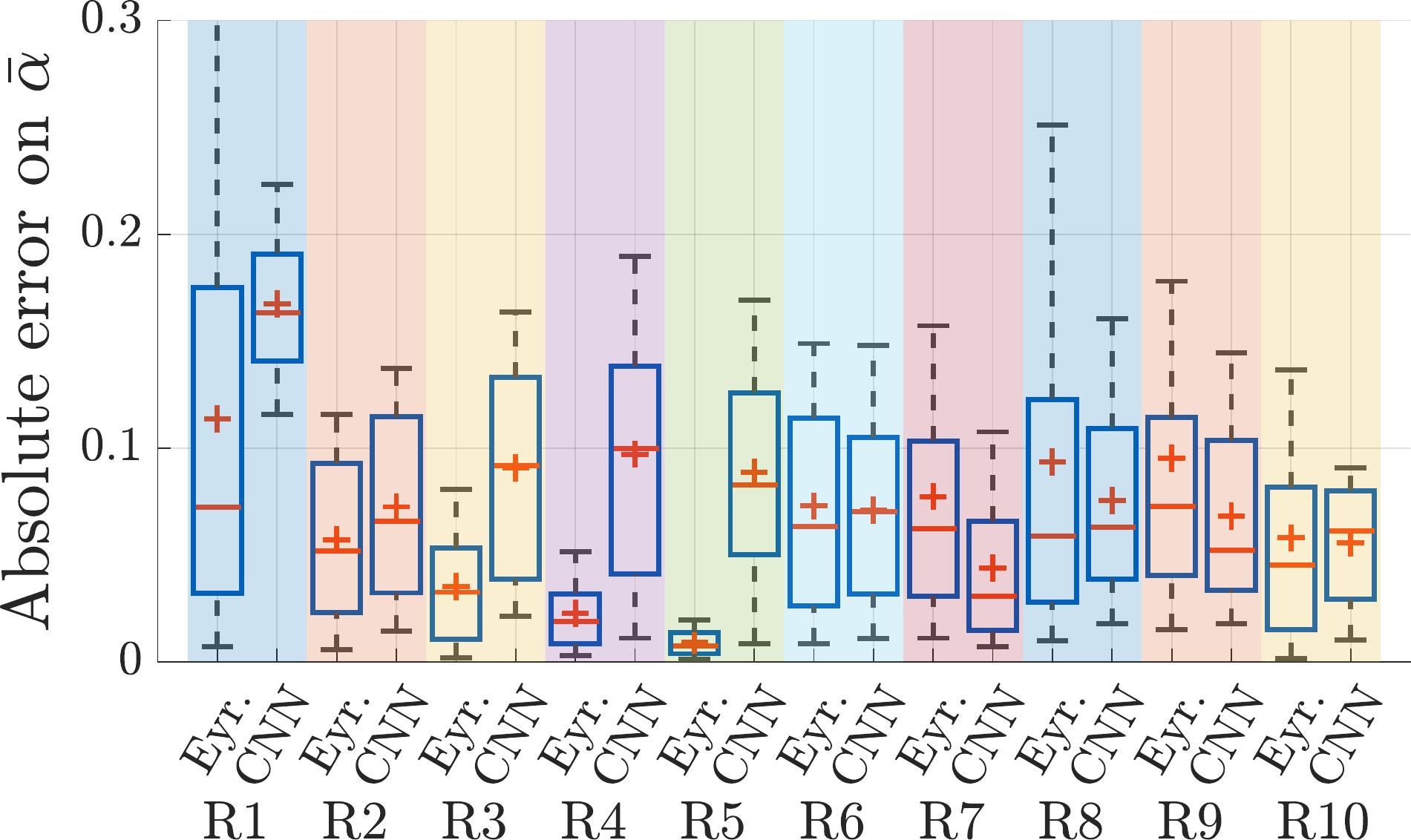}
		\caption{Comparison of $\bar{\alpha}(\textrm{1000 Hz})$ estimation errors over measured RIRs in 10 rooms using Eyring and CNN-RB. Only selected RIRs with Schroeder curves in $\mathcal{A}$ are included.}
		\label{fig:real_1000}
	\end{minipage}
	\hfill
	\begin{minipage}{0.39\textwidth}
		\includegraphics[height=0.25\textheight]{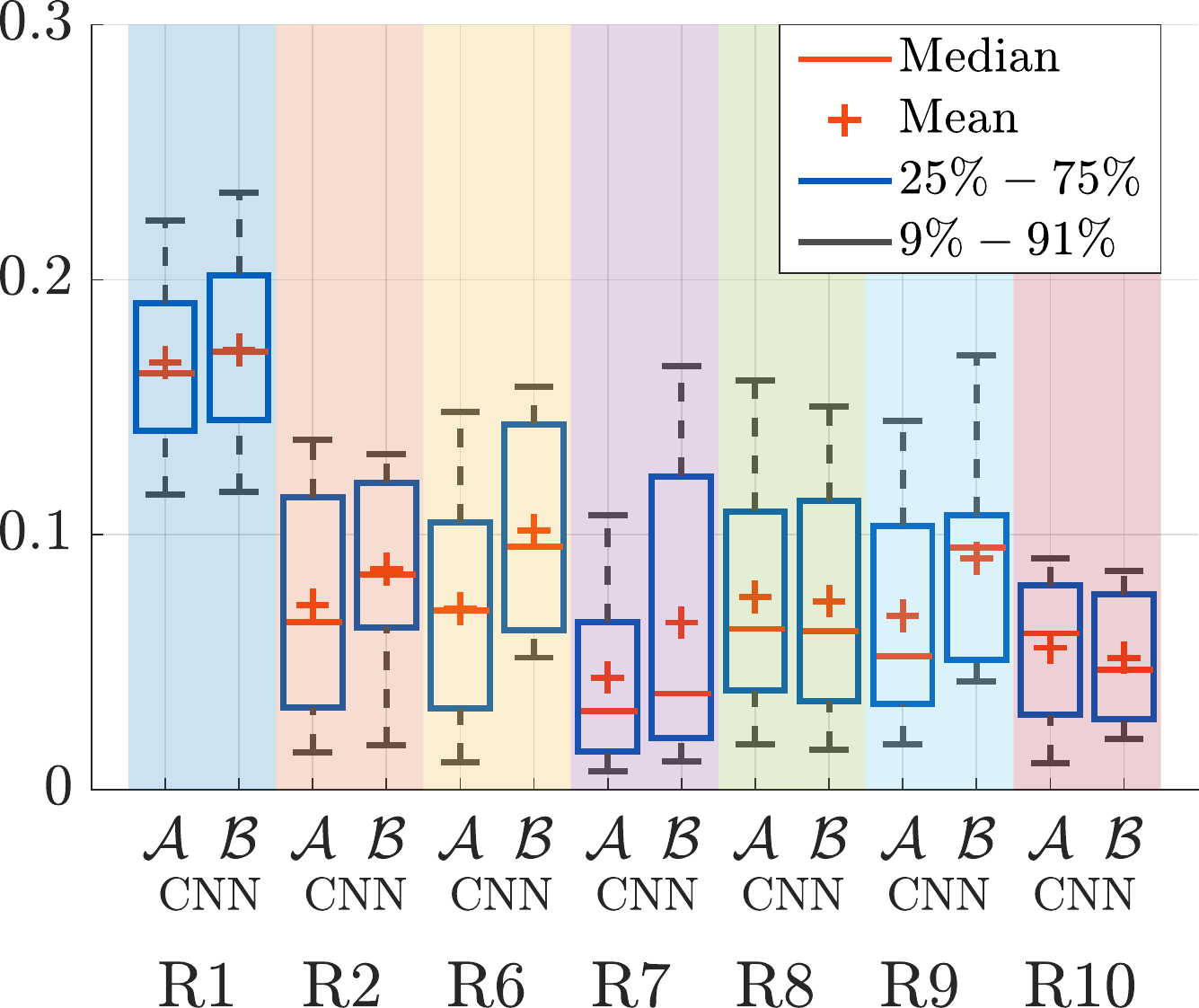}
		\caption{Comparison of $\bar{\alpha}(\textrm{1000 Hz})$ CNN-RB estimation errors over measured RIRs with 1~kHz Schroeder curves in $\mathcal{A}$ vs. in $\mathcal{B}$.}
		\label{fig:real_1000_AB}
	\end{minipage}
\end{figure*}

The right part of Fig.~\ref{fig:real_results_all} reports mean and stds of absolute errors for the CNN-RB model, using only the RIRs in $\mathcal{A}$.
%
\note{\sout{Encouragingly, for the 1~kHz, 2~kHz and 4~kHz octave bands, the learning-based method yields errors comparable to Eyring's formula in almost all rooms, except the three most reverberant rooms (R3, R4 and R5) in which the latter performs very well.
		Means and stds of errors from both methods are below or around 0.1 for all rooms, which is a reasonable uncertainty in the context of acoustic diagnosis. However, CNN errors in the 500~Hz octave band are very large in all rooms except R1 and R8. This could be due to unforeseen discrepancies between simulated and real RIRs in this band, and needs further investigation. Fig.~\ref{fig:real_1000} shows the same results in the form of bar plots for the 1~kHz octave band only, further confirming that the CNN yields error distribution equivalent to Eyring's in most rooms.}}
%
%
\textcolor{black}{Encouragingly, for the 1~kHz, 2~kHz and 4~kHz octave bands, the learning-based method yields errors below or around 0.1 for all rooms, which is a reasonable uncertainty in the context of acoustic diagnosis.}
Errors are comparable to Eyring's formula except in the three most reverberant ones (R3, R4 and R5) for which the latter performs very well.
\textcolor{black}{For the octave band centered at 4~kz, the CNN-RB errors increase slightly. A possible explanation could lie in the stronger directivity of the source at this frequency, as observed in the manufacturer's data (recall that the neural network has only been trained on omnidirectional sources).  For the octave band centered at 500 Hz, the CNN-RB errors are much larger in all rooms except R1 and R8. One of the preferred hypotheses is the existence of a wave phenomenon in this band that could not be learned by the neural network trained on Roomsim.
	These hypotheses will need to be validated by further research on real data. }
Fig.~\ref{fig:real_1000} shows the same results in the form of bar plots for the 1~kHz octave band, further confirming that the CNN-RB model yields error distributions comparable to Eyring's in this band.

Finally, Fig.~\ref{fig:real_1000_AB} compares errors obtained with the CNN-RB on measured RIRs whose 1~kHz Schroeder curves are in $\mathcal{A}$ against those whose Schroeder curves are in $\mathcal{B}$. Note that rooms R3, R4 and R5 are omitted here because an insufficient number of curves were placed in $\mathcal{B}$ for these rooms. Encouragingly, we observe that the CNN is largely unaffected by the non-linear or insufficient log-energy decays of Schroeder curves in $\mathcal{B}$. This suggests that the network learned to rely on more elaborate and more robust features than those used by reverberation-based techniques. In contrast, obtaining reliable absorption estimates from these curves using Eyring's model was fundamentally impossible, due to its reliance on reverberation time.

\section{\label{sec:conclusion} Conclusion}
\textcolor{black}{In this work, we tackled the inverse problem of estimating the area-weighted mean absorption coefficients of a room from a single RIR using virtually-supervized learning, in a broad range of acoustical conditions pertaining to the field of building acoustic diagnosis. Different neural network designs and simulated training strategies were proposed, explored and tested. The developed methods were compared to classical formulas that hinge on the room's volume, total surface, reverberation time and on the diffuse sound field (DSF) hypothesis. In close-to-DSF conditions, our experiments on both simulated and real data revealed that the best learned models yielded estimation errors comparable to classical ones without needing the room's geometry. As expected and predicted by reverberation theory, the performances of DSF-based models degraded under conditions departing from DSF. These include rooms featuring less reverberation, less diffusion, non-homogenous geometries, and more generally RIRs featuring insufficient or non-linear decays of their Schroeder curves. In contrast, the proposed virtually-trained models showed remarkable robustness in estimating the target quantity under such conditions, suggesting that they learned to rely on more elaborate and more robust features than those used by reverberation-based techniques.}


This first extensive experimental study on virtually-supervised mean absorption estimation aimed at paving the way towards simpler and more robust acoustic diagnosis techniques. Future works will include \textcolor{black}{further experimental investigations on the poorer performance of the learned models at lower frequencies on real data, notably by employing higher-end sound sources}. Leads for improving the learned models include domain adaption, data augmentation and probabilistic uncertainty modeling. We also plan to build on our findings to tackle the much more difficult problem of estimating the absorption coefficients of \textit{individual} surfaces from RIRs. \textcolor{black}{For this, geometrically-informed models and the aggregation of RIRs from multiple source-receiver pairs will be leveraged.}



\begin{thebibliography}{0}
\def\enquote#1{``#1,''}
\def\plainquote#1{``#1''}
\expandafter\ifx\csname natexlab\endcsname\relax\def\natexlab#1{#1}\fi
\providecommand{\dourl}[1]{\href{http://#1}{\nolinkurl{#1}}}
\providecommand{\bibinfo}[2]{#2}
\providecommand{\noopsort}[1]{}
\providecommand{\switchargs}[2]{#2#1}
  \def\eatspace #1{#1}

\end{thebibliography}


\begin{thebibliography}{80}
	\def\enquote#1{``#1,''}
	\def\plainquote#1{``#1''}
	\expandafter\ifx\csname natexlab\endcsname\relax\def\natexlab#1{#1}\fi
	\providecommand{\dourl}[1]{\href{http://#1}{\nolinkurl{#1}}}
	\providecommand{\bibinfo}[2]{#2}
	\providecommand{\noopsort}[1]{}
	\providecommand{\switchargs}[2]{#2#1}
	\def\eatspace #1{#1}
	
	\bibitem[{Allard and Aknine(1985)}]{Allard1985b}
	\bibinfo{author}{Allard, J.~F.},  and \bibinfo{author}{Aknine, A.}
	(\textbf{\bibinfo{year}{1985}}). \enquote{\bibinfo{title}{Acoustic impedance
			measurements with a sound intensity meter}} \bibinfo{journal}{Applied
		Acoustics} \textbf{18}, \bibinfo{pages}{69--75}.
	
	\bibitem[{Allard and Sieben(1985)}]{Allard1985a}
	\bibinfo{author}{Allard, J.~F.},  and \bibinfo{author}{Sieben, B.}
	(\textbf{\bibinfo{year}{1985}}). \enquote{\bibinfo{title}{Measurements of
			acoustic impedance in a free field with two microphones and a spectrum
			analyzer}} \bibinfo{journal}{The Journal of the Acoustical Society of
		America} \textbf{77}, \bibinfo{pages}{1617--1618}.
	
	\bibitem[{Allen and Berkley(1979)}]{allen1979image}
	\bibinfo{author}{Allen, J.~B.},  and \bibinfo{author}{Berkley, D.~A.}
	(\textbf{\bibinfo{year}{1979}}). \enquote{\bibinfo{title}{Image method for
			efficiently simulating small-room acoustics}} \bibinfo{journal}{The Journal
		of the Acoustical Society of America} \textbf{65}(4),
	\bibinfo{pages}{943--950}.
	
	\bibitem[{Ando(1968)}]{Ando1968}
	\bibinfo{author}{Ando, Y.} (\textbf{\bibinfo{year}{1968}}).
	\enquote{\bibinfo{title}{The interference pattern method of measuring the
			complex reflection coefficient of acoustic materials at oblique incidence}}
	in \emph{\bibinfo{booktitle}{Proc. 6 th International Congress on
			Acoustics}}, \bibinfo{organization}{ICA}.
	
	\bibitem[{Aoshima(1981)}]{Aoshima1981}
	\bibinfo{author}{Aoshima, N.} (\textbf{\bibinfo{year}{1981}}).
	\enquote{\bibinfo{title}{Computer-generated pulse signal applied for sound
			measurement}} \bibinfo{journal}{The Journal of the Acoustical Society of
		America} \textbf{69}, \bibinfo{pages}{1484--1488}.
	
	\bibitem[{ASTM E1050-98()}]{ASTME1050}
	ASTM E1050-98 (\textbf{\bibinfo{year}{2006}}).
	\enquote{\bibinfo{title}{Standard test method for impedance and absorption of
			acoustical materials using a tube, two microphones, and a digital frequency
			analysis system}} \bibinfo{type}{Standard}.
	
	\bibitem[{Barry(1974)}]{Barry1974}
	\bibinfo{author}{Barry, T.} (\textbf{\bibinfo{year}{1974}}).
	\enquote{\bibinfo{title}{Measurement of the absorption spectrum using
			correlation / spectral density techniques}} \bibinfo{journal}{The Journal of
		the Acoustical Society of America} \textbf{55}, \bibinfo{pages}{1349--1351}.
	
	\bibitem[{Bianco \emph{et~al.}(2019)Bianco, Gerstoft, Traer, Ozanich, Roch,
		Gannot, and Deledalle}]{bianco2019machine}
	\bibinfo{author}{Bianco, M.~J.}, \bibinfo{author}{Gerstoft, P.},
	\bibinfo{author}{Traer, J.}, \bibinfo{author}{Ozanich, E.},
	\bibinfo{author}{Roch, M.~A.}, \bibinfo{author}{Gannot, S.},  and
	\bibinfo{author}{Deledalle, C.-A.} (\textbf{\bibinfo{year}{2019}}).
	\enquote{\bibinfo{title}{Machine learning in acoustics: Theory and
			applications}} \bibinfo{journal}{The Journal of the Acoustical Society of
		America} \textbf{146}(5), \bibinfo{pages}{3590--3628}.
	
	\bibitem[{Borish(1984)}]{borish1984extension}
	\bibinfo{author}{Borish, J.} (\textbf{\bibinfo{year}{1984}}).
	\enquote{\bibinfo{title}{Extension of the image model to arbitrary
			polyhedra}} \bibinfo{journal}{The Journal of the Acoustical Society of
		America} \textbf{75}(6), \bibinfo{pages}{1827--1836}.
	
	\bibitem[{Botteldooren(1995)}]{botteldooren1995finite}
	\bibinfo{author}{Botteldooren, D.} (\textbf{\bibinfo{year}{1995}}).
	\enquote{\bibinfo{title}{Finite-difference time-domain simulation of
			low-frequency room acoustic problems}} \bibinfo{journal}{The Journal of the
		Acoustical Society of America} \textbf{98}(6), \bibinfo{pages}{3302--3308}.
	
	\bibitem[{Brandão \emph{et~al.}(2015)Brandão, Lenzi, and Paul}]{Brandao2015l}
	\bibinfo{author}{Brandão, E.}, \bibinfo{author}{Lenzi, A.},  and
	\bibinfo{author}{Paul, S.} (\textbf{\bibinfo{year}{2015}}).
	\enquote{\bibinfo{title}{A review of the in situ impedance and sound
			absorption measurement techniques}} \bibinfo{journal}{Acta Acustica United
		with Acustica} \textbf{101}(3), \bibinfo{pages}{443--463}.
	
	\bibitem[{Chakrabarty and Habets(2017)}]{chakrabarty2017broadband}
	\bibinfo{author}{Chakrabarty, S.},  and \bibinfo{author}{Habets, E.~A.}
	(\textbf{\bibinfo{year}{2017}}). \enquote{\bibinfo{title}{Broadband doa
			estimation using convolutional neural networks trained with noise signals}}
	in \emph{\bibinfo{booktitle}{2017 IEEE Workshop on Applications of Signal
			Processing to Audio and Acoustics (WASPAA)}}, \bibinfo{organization}{IEEE},
	pp. \bibinfo{pages}{136--140}.
	
	\bibitem[{Champoux and L’espérance(1988)}]{Champoux1988b}
	\bibinfo{author}{Champoux, Y.},  and \bibinfo{author}{L’espérance, A.}
	(\textbf{\bibinfo{year}{1988}}). \enquote{\bibinfo{title}{Numerical
			evaluation of errors associated with the measurement of acoustic impedance in
			a free field using two microphones and a spectrum analyzer}}
	\bibinfo{journal}{The Journal of the Acoustical Society of America}
	\textbf{84}, \bibinfo{pages}{30--38}.
	
	\bibitem[{Champoux \emph{et~al.}(1988)Champoux, Nicolas, and
		Allard}]{Champoux1988a}
	\bibinfo{author}{Champoux, Y.}, \bibinfo{author}{Nicolas, J.},  and
	\bibinfo{author}{Allard, J.~F.} (\textbf{\bibinfo{year}{1988}}).
	\enquote{\bibinfo{title}{Measurement of acoustic impedance in a free field at
			low frequencies}} \bibinfo{journal}{Journal of Sound and Vibration}
	\textbf{125}, \bibinfo{pages}{313--323}.
	
	\bibitem[{Chung and Blaser(1980{\natexlab{a}})}]{Chung1980a}
	\bibinfo{author}{Chung, J.},  and \bibinfo{author}{Blaser, D.}
	(\textbf{\bibinfo{year}{1980}}{\natexlab{a}}).
	\enquote{\bibinfo{title}{Transfer function method of measuring in-duct
			acoustic properties. i. theory}} \bibinfo{journal}{The Journal of the
		Acoustical Society of America} \textbf{68}(3), \bibinfo{pages}{907--913}.
	
	\bibitem[{Chung and Blaser(1980{\natexlab{b}})}]{Chung1980b}
	\bibinfo{author}{Chung, J.},  and \bibinfo{author}{Blaser, D.}
	(\textbf{\bibinfo{year}{1980}}{\natexlab{b}}).
	\enquote{\bibinfo{title}{Transfer function method of measuring in-duct
			acoustic properties. ii: Experiment}} \bibinfo{journal}{The Journal of the
		Acoustical Society of America} \textbf{68}(3), \bibinfo{pages}{914--921}.
	
	\bibitem[{Cramond(1984)}]{Cramond1984}
	\bibinfo{author}{Cramond, A. J.and~Don, C.~G.} (\textbf{\bibinfo{year}{1984}}).
	\enquote{\bibinfo{title}{Reflection of impulses as a method of determining
			acoustic impedance}} \bibinfo{journal}{The Journal of the Acoustical Society
		of America} \textbf{75}, \bibinfo{pages}{382--389}.
	
	\bibitem[{Davies and Mulholland(1979)}]{Davies1979}
	\bibinfo{author}{Davies, J.~C.},  and \bibinfo{author}{Mulholland, K.~A.}
	(\textbf{\bibinfo{year}{1979}}). \enquote{\bibinfo{title}{An impulse method
			of measuring normal impedance at oblique incidence}}
	\bibinfo{journal}{Journal of Sound and Vibration} \textbf{67},
	\bibinfo{pages}{135--149}.
	
	\bibitem[{Deecke and Janik(2006)}]{deecke2006automated}
	\bibinfo{author}{Deecke, V.~B.},  and \bibinfo{author}{Janik, V.~M.}
	(\textbf{\bibinfo{year}{2006}}). \enquote{\bibinfo{title}{Automated
			categorization of bioacoustic signals: avoiding perceptual pitfalls}}
	\bibinfo{journal}{The Journal of the Acoustical Society of America}
	\textbf{119}(1), \bibinfo{pages}{645--653}.
	
	\bibitem[{Deleforge \emph{et~al.}(2014)Deleforge, Forbes, and
		Horaud}]{deleforge2014acoustic}
	\bibinfo{author}{Deleforge, A.}, \bibinfo{author}{Forbes, F.},  and
	\bibinfo{author}{Horaud, R.} (\textbf{\bibinfo{year}{2014}}).
	\enquote{\bibinfo{title}{Acoustic space learning for sound-source separation
			and localization on binaural manifolds}} \bibinfo{journal}{International
		journal of neural systems} \textbf{25}(01), \bibinfo{pages}{1440003}.
	
	\bibitem[{Deleforge \emph{et~al.}(2015)Deleforge, Horaud, Schechner, and
		Girin}]{deleforge2015co}
	\bibinfo{author}{Deleforge, A.}, \bibinfo{author}{Horaud, R.},
	\bibinfo{author}{Schechner, Y.~Y.},  and \bibinfo{author}{Girin, L.}
	(\textbf{\bibinfo{year}{2015}}). \enquote{\bibinfo{title}{Co-localization of
			audio sources in images using binaural features and locally-linear
			regression}} \bibinfo{journal}{IEEE/ACM Transactions on Audio, Speech, and
		Language Processing} \textbf{23}(4), \bibinfo{pages}{718--731}.
	
	\bibitem[{Di~Carlo \emph{et~al.}(2019)Di~Carlo, Deleforge, and
		Bertin}]{di2019mirage}
	\bibinfo{author}{Di~Carlo, D.}, \bibinfo{author}{Deleforge, A.},  and
	\bibinfo{author}{Bertin, N.} (\textbf{\bibinfo{year}{2019}}).
	\enquote{\bibinfo{title}{{MIRAGE}: {2D} source localization using microphone
			pair augmentation with echoes}} in \emph{\bibinfo{booktitle}{ICASSP 2019-2019
			IEEE International Conference on Acoustics, Speech and Signal Processing
			(ICASSP)}}, \bibinfo{organization}{IEEE}, pp. \bibinfo{pages}{775--779}.
	
	\bibitem[{Di~Carlo \emph{et~al.}(2021)Di~Carlo, Tandeitnik, Foy, Deleforge,
		Bertin, and Gannot}]{di2021dechorate}
	\bibinfo{author}{Di~Carlo, D.}, \bibinfo{author}{Tandeitnik, P.},
	\bibinfo{author}{Foy, C.}, \bibinfo{author}{Deleforge, A.},
	\bibinfo{author}{Bertin, N.},  and \bibinfo{author}{Gannot, S.}
	(\textbf{\bibinfo{year}{2021}}). \enquote{\bibinfo{title}{dechorate: a
			calibrated room impulse response database for echo-aware signal processing}}
	\bibinfo{journal}{arXiv preprint arXiv:2104.13168} .
	
	\bibitem[{Farina(2000)}]{farina2000}
	\bibinfo{author}{Farina, A.} (\textbf{\bibinfo{year}{2000}}).
	\enquote{\bibinfo{title}{Simultaneous measurement of impulse response and
			distortion with a swept-sine technique}} in \emph{\bibinfo{booktitle}{Audio
			Engineering Society Convention 108}}, \bibinfo{organization}{Audio
		Engineering Society}.
	
	\bibitem[{Farina(2007)}]{farina2007advancements}
	\bibinfo{author}{Farina, A.} (\textbf{\bibinfo{year}{2007}}).
	\enquote{\bibinfo{title}{Advancements in impulse response measurements by
			sine sweeps}} in \emph{\bibinfo{booktitle}{Audio Engineering Society
			Convention 122}}, \bibinfo{organization}{Audio Engineering Society}.
	
	\bibitem[{Gamper and Tashev(2018)}]{gamper2018blind}
	\bibinfo{author}{Gamper, H.},  and \bibinfo{author}{Tashev, I.~J.}
	(\textbf{\bibinfo{year}{2018}}). \enquote{\bibinfo{title}{Blind reverberation
			time estimation using a convolutional neural network}} in
	\emph{\bibinfo{booktitle}{2018 16th International Workshop on Acoustic Signal
			Enhancement (IWAENC)}}, \bibinfo{organization}{IEEE}, pp.
	\bibinfo{pages}{136--140}.
	
	\bibitem[{Garai(1993)}]{Garai1993}
	\bibinfo{author}{Garai, M.} (\textbf{\bibinfo{year}{1993}}).
	\enquote{\bibinfo{title}{Measurement of the sound-absorption coefficient in
			situ: the reflection method using periodic pseudorandom sequences of maximum
			length}} \bibinfo{journal}{Applied Acoustics} \textbf{39},
	\bibinfo{pages}{119--139}.
	
	\bibitem[{Gaultier \emph{et~al.}(2017)Gaultier, Kataria, and
		Deleforge}]{gaultier2017vast}
	\bibinfo{author}{Gaultier, C.}, \bibinfo{author}{Kataria, S.},  and
	\bibinfo{author}{Deleforge, A.} (\textbf{\bibinfo{year}{2017}}).
	\enquote{\bibinfo{title}{Vast: The virtual acoustic space traveler dataset}}
	in \emph{\bibinfo{booktitle}{International Conference on Latent Variable
			Analysis and Signal Separation}}, \bibinfo{organization}{Springer}, pp.
	\bibinfo{pages}{68--79}.
	
	\bibitem[{Genovese \emph{et~al.}(2019)Genovese, Gamper, Pulkki, Raghuvanshi,
		and Tashev}]{genovese2019blind}
	\bibinfo{author}{Genovese, A.~F.}, \bibinfo{author}{Gamper, H.},
	\bibinfo{author}{Pulkki, V.}, \bibinfo{author}{Raghuvanshi, N.},  and
	\bibinfo{author}{Tashev, I.~J.} (\textbf{\bibinfo{year}{2019}}).
	\enquote{\bibinfo{title}{Blind room volume estimation from single-channel
			noisy speech}} in \emph{\bibinfo{booktitle}{ICASSP 2019-2019 IEEE
			International Conference on Acoustics, Speech and Signal Processing
			(ICASSP)}}, \bibinfo{organization}{IEEE}, pp. \bibinfo{pages}{231--235}.
	
	\bibitem[{Gradi{\v{s}}ek \emph{et~al.}(2017)Gradi{\v{s}}ek, Slapni{\v{c}}ar,
		{\v{S}}orn, Lu{\v{s}}trek, Gams, and Grad}]{gradivsek2017predicting}
	\bibinfo{author}{Gradi{\v{s}}ek, A.}, \bibinfo{author}{Slapni{\v{c}}ar, G.},
	\bibinfo{author}{{\v{S}}orn, J.}, \bibinfo{author}{Lu{\v{s}}trek, M.},
	\bibinfo{author}{Gams, M.},  and \bibinfo{author}{Grad, J.}
	(\textbf{\bibinfo{year}{2017}}). \enquote{\bibinfo{title}{Predicting species
			identity of bumblebees through analysis of flight buzzing sounds}}
	\bibinfo{journal}{Bioacoustics} \textbf{26}(1), \bibinfo{pages}{63--76}.
	
	\bibitem[{Guidorzia \emph{et~al.}(2015)Guidorzia, Barbaresia, D’Orazioa, and
		Garai}]{Guidorzia2015}
	\bibinfo{author}{Guidorzia, P.}, \bibinfo{author}{Barbaresia, L.},
	\bibinfo{author}{D’Orazioa, D.},  and \bibinfo{author}{Garai, M.}
	(\textbf{\bibinfo{year}{2015}}). \enquote{\bibinfo{title}{Impulse responses
			measured with mls or swept-sine signals applied to architectural acoustics:
			an in-depth analysis of the two methods and some case studies of measurements
			inside theaters}} in \emph{\bibinfo{booktitle}{6 th International Building
			Physics Conference}}, \bibinfo{organization}{IBPC}.
	
	\bibitem[{Habets(2006)}]{habets2006room}
	\bibinfo{author}{Habets, E.~A.} (\textbf{\bibinfo{year}{2006}}).
	\enquote{\bibinfo{title}{Room impulse response generator}}
	\bibinfo{journal}{Technische Universiteit Eindhoven, Tech. Rep}
	\textbf{2}(2.4), \bibinfo{pages}{1}.
	
	\bibitem[{He \emph{et~al.}(2019)He, Motlicek, and Odobez}]{he2019adaptation}
	\bibinfo{author}{He, W.}, \bibinfo{author}{Motlicek, P.},  and
	\bibinfo{author}{Odobez, J.-M.} (\textbf{\bibinfo{year}{2019}}).
	\enquote{\bibinfo{title}{Adaptation of multiple sound source localization
			neural networks with weak supervision and domain-adversarial training}} in
	\emph{\bibinfo{booktitle}{ICASSP 2019-2019 IEEE International Conference on
			Acoustics, Speech and Signal Processing (ICASSP)}},
	\bibinfo{organization}{IEEE}, pp. \bibinfo{pages}{770--774}.
	
	\bibitem[{Hodgson(1994)}]{Hodgson94}
	\bibinfo{author}{Hodgson, M.} (\textbf{\bibinfo{year}{1994}}).
	\enquote{\bibinfo{title}{When is diffuse-field theory accurate?}}
	\bibinfo{journal}{Canadian Acoustics - Acoustique Canadienne} \textbf{22}(3),
	\bibinfo{pages}{41--42}.
	
	\bibitem[{Hodgson(1996)}]{Hodgson96}
	\bibinfo{author}{Hodgson, M.} (\textbf{\bibinfo{year}{1996}}).
	\enquote{\bibinfo{title}{When is diffuse-field theory applicable?}}
	\bibinfo{journal}{Applied Acoustics} \textbf{49}(3),
	\bibinfo{pages}{191--201}.
	
	\bibitem[{Hollin and Jones(1977)}]{Hollin1977l}
	\bibinfo{author}{Hollin, K.~A.},  and \bibinfo{author}{Jones, M.~H.}
	(\textbf{\bibinfo{year}{1977}}). \enquote{\bibinfo{title}{The measurement of
			sound absorption coefficient in situ by a correlation technique}}
	\bibinfo{journal}{Acustica} \textbf{37}, \bibinfo{pages}{103--110}.
	
	\bibitem[{Ingård and Bolt(1951)}]{Ingard1951}
	\bibinfo{author}{Ingård, U.},  and \bibinfo{author}{Bolt, R.~H.}
	(\textbf{\bibinfo{year}{1951}}). \enquote{\bibinfo{title}{A free field method
			of measuring the absorption coefficient of acoustic materials}}
	\bibinfo{journal}{The Journal of the Acoustical Society of America}
	\textbf{23}, \bibinfo{pages}{509--516}.
	
	\bibitem[{ISO 10534:2001()}]{ISO10534}
	ISO 10534:2001 (\textbf{\bibinfo{year}{2001}}).
	\enquote{\bibinfo{title}{Acoustics. determination of sound absorption
			coefficient and impedance in impedance tubes. part 1: Method using standing
			wave part 2: Transfer function method}} \bibinfo{type}{Standard}.
	
	\bibitem[{ISO 3382-2:2008()}]{ISO3382}
	ISO 3382-2:2008 (\textbf{\bibinfo{year}{2008}}).
	\enquote{\bibinfo{title}{Acoustics — measurement of room acoustic
			parameters — part 2: Reverberation time in ordinary rooms}}
	\bibinfo{type}{Standard}.
	
	\bibitem[{ISO 354:2003()}]{ISO354}
	ISO 354:2003 (\textbf{\bibinfo{year}{2003}}).
	\enquote{\bibinfo{title}{Acoustics - measurement of sound absorption in a
			reverberation room}} \bibinfo{type}{Standard}.
	
	\bibitem[{Kataria \emph{et~al.}(2017)Kataria, Gaultier, and
		Deleforge}]{kataria2017hearing}
	\bibinfo{author}{Kataria, S.}, \bibinfo{author}{Gaultier, C.},  and
	\bibinfo{author}{Deleforge, A.} (\textbf{\bibinfo{year}{2017}}).
	\enquote{\bibinfo{title}{Hearing in a shoe-box: binaural source position and
			wall absorption estimation using virtually supervised learning}} in
	\emph{\bibinfo{booktitle}{2017 IEEE International Conference on Acoustics,
			Speech and Signal Processing (ICASSP)}}, \bibinfo{organization}{IEEE}, pp.
	\bibinfo{pages}{226--230}.
	
	\bibitem[{Kim \emph{et~al.}(2017)Kim, Misra, Chin, Hughes, Narayanan, Sainath,
		and Bacchiani}]{kim2017generation}
	\bibinfo{author}{Kim, C.}, \bibinfo{author}{Misra, A.}, \bibinfo{author}{Chin,
		K.}, \bibinfo{author}{Hughes, T.}, \bibinfo{author}{Narayanan, A.},
	\bibinfo{author}{Sainath, T.},  and \bibinfo{author}{Bacchiani, M.}
	(\textbf{\bibinfo{year}{2017}}). \enquote{\bibinfo{title}{Generation of
			large-scale simulated utterances in virtual rooms to train deep-neural
			networks for far-field speech recognition in google home}} .
	
	\bibitem[{Kingma and Ba(2014)}]{kingma2014adam}
	\bibinfo{author}{Kingma, D.~P.},  and \bibinfo{author}{Ba, J.}
	(\textbf{\bibinfo{year}{2014}}). \enquote{\bibinfo{title}{Adam: A method for
			stochastic optimization}} \bibinfo{journal}{arXiv preprint arXiv:1412.6980} .
	
	\bibitem[{Kulowski(1985)}]{kulowski1985algorithmic}
	\bibinfo{author}{Kulowski, A.} (\textbf{\bibinfo{year}{1985}}).
	\enquote{\bibinfo{title}{Algorithmic representation of the ray tracing
			technique}} \bibinfo{journal}{Applied Acoustics} \textbf{18}(6),
	\bibinfo{pages}{449--469}.
	
	\bibitem[{Kuttruff(2009)}]{Kuttruff:09}
	\bibinfo{author}{Kuttruff, H.} (\textbf{\bibinfo{year}{2009}}).
	\emph{\bibinfo{title}{Room Acoustics - Fifth edition}}
	(\bibinfo{publisher}{Spon Press}, \bibinfo{address}{Oxfordshire, England}).
	
	\bibitem[{Lefort \emph{et~al.}(2017)Lefort, Real, and
		Dr{\'e}meau}]{lefort2017direct}
	\bibinfo{author}{Lefort, R.}, \bibinfo{author}{Real, G.},  and
	\bibinfo{author}{Dr{\'e}meau, A.} (\textbf{\bibinfo{year}{2017}}).
	\enquote{\bibinfo{title}{Direct regressions for underwater acoustic source
			localization in fluctuating oceans}} \bibinfo{journal}{Applied Acoustics}
	\textbf{116}, \bibinfo{pages}{303--310}.
	
	\bibitem[{Li and Hodgson(1997)}]{Li1997}
	\bibinfo{author}{Li, J.~F.},  and \bibinfo{author}{Hodgson, M.}
	(\textbf{\bibinfo{year}{1997}}). \enquote{\bibinfo{title}{Use of
			pseudo-random sequences and a single microphone to measure surface impedance
			at oblique incidence}} \bibinfo{journal}{The Journal of the Acoustical
		Society of America} \textbf{102}, \bibinfo{pages}{2200--2210}.
	
	\bibitem[{Luo and Mesgarani(2018)}]{luo2018tasnet}
	\bibinfo{author}{Luo, Y.},  and \bibinfo{author}{Mesgarani, N.}
	(\textbf{\bibinfo{year}{2018}}). \enquote{\bibinfo{title}{Tasnet: time-domain
			audio separation network for real-time, single-channel speech separation}} in
	\emph{\bibinfo{booktitle}{2018 IEEE International Conference on Acoustics,
			Speech and Signal Processing (ICASSP)}}, \bibinfo{organization}{IEEE}, pp.
	\bibinfo{pages}{696--700}.
	
	\bibitem[{Mesaros \emph{et~al.}(2017)Mesaros, Heittola, Diment, Elizalde, Shah,
		Vincent, Raj, and Virtanen}]{mesaros2017dcase}
	\bibinfo{author}{Mesaros, A.}, \bibinfo{author}{Heittola, T.},
	\bibinfo{author}{Diment, A.}, \bibinfo{author}{Elizalde, B.},
	\bibinfo{author}{Shah, A.}, \bibinfo{author}{Vincent, E.},
	\bibinfo{author}{Raj, B.},  and \bibinfo{author}{Virtanen, T.}
	(\textbf{\bibinfo{year}{2017}}). \enquote{\bibinfo{title}{Dcase 2017
			challenge setup: Tasks, datasets and baseline system}} in
	\emph{\bibinfo{booktitle}{DCASE 2017-Workshop on Detection and Classification
			of Acoustic Scenes and Events}}.
	
	\bibitem[{Mesaros \emph{et~al.}(2019)Mesaros, Heittola, and
		Virtanen}]{mesaros2019acoustic}
	\bibinfo{author}{Mesaros, A.}, \bibinfo{author}{Heittola, T.},  and
	\bibinfo{author}{Virtanen, T.} (\textbf{\bibinfo{year}{2019}}).
	\enquote{\bibinfo{title}{Acoustic scene classification in dcase 2019
			challenge: Closed and open set classification and data mismatch setups}} .
	
	\bibitem[{Minten \emph{et~al.}(1988)Minten, Cops, and Lauriks}]{Minten1988}
	\bibinfo{author}{Minten, M.}, \bibinfo{author}{Cops, A.},  and
	\bibinfo{author}{Lauriks, W.} (\textbf{\bibinfo{year}{1988}}).
	\enquote{\bibinfo{title}{Absorption characteristics of an acoustic material
			at oblique incidence measured with the two-microphone technique}}
	\bibinfo{journal}{Journal of Sound and Vibration} \textbf{120},
	\bibinfo{pages}{499--510}.
	
	\bibitem[{Müller and Massarani(2001)}]{Muller2001}
	\bibinfo{author}{Müller, S.},  and \bibinfo{author}{Massarani, P.}
	(\textbf{\bibinfo{year}{2001}}). \enquote{\bibinfo{title}{Transfer-function
			measurement with sweeps. director's cut including previously unreleased
			material and some corrections}} \bibinfo{journal}{Journal of the Audio
		Engineering Society} \textbf{49}(6), \bibinfo{pages}{443--471}.
	
	\bibitem[{Niu \emph{et~al.}(2017)Niu, Reeves, and Gerstoft}]{niu2017source}
	\bibinfo{author}{Niu, H.}, \bibinfo{author}{Reeves, E.},  and
	\bibinfo{author}{Gerstoft, P.} (\textbf{\bibinfo{year}{2017}}).
	\enquote{\bibinfo{title}{Source localization in an ocean waveguide using
			supervised machine learning}} \bibinfo{journal}{The Journal of the Acoustical
		Society of America} \textbf{142}(3), \bibinfo{pages}{1176--1188}.
	
	\bibitem[{Nobile and Hayek(1985)}]{Nobile1985}
	\bibinfo{author}{Nobile, M.~A.},  and \bibinfo{author}{Hayek, S.~I.}
	(\textbf{\bibinfo{year}{1985}}). \enquote{\bibinfo{title}{Acoustic
			propagation over an impedance plane}} \bibinfo{journal}{The Journal of the
		Acoustical Society of America} \textbf{78}, \bibinfo{pages}{1325--1336}.
	
	\bibitem[{Nolan(2020)}]{nolan2020estimation}
	\bibinfo{author}{Nolan, M.} (\textbf{\bibinfo{year}{2020}}).
	\enquote{\bibinfo{title}{Estimation of angle-dependent absorption
			coefficients from spatially distributed in situ measurements}}
	\bibinfo{journal}{The Journal of the Acoustical Society of America}
	\textbf{147}(2), \bibinfo{pages}{EL119--EL124}.
	
	\bibitem[{Nolan \emph{et~al.}(2018)Nolan, Fernandez-Grande, Brunskog, and
		Jeong}]{nolan2018wavenumber}
	\bibinfo{author}{Nolan, M.}, \bibinfo{author}{Fernandez-Grande, E.},
	\bibinfo{author}{Brunskog, J.},  and \bibinfo{author}{Jeong, C.-H.}
	(\textbf{\bibinfo{year}{2018}}). \enquote{\bibinfo{title}{A wavenumber
			approach to quantifying the isotropy of the sound field in reverberant
			spaces}} \bibinfo{journal}{The Journal of the Acoustical Society of America}
	\textbf{143}(4), \bibinfo{pages}{2514--2526}.
	
	\bibitem[{Okuzono \emph{et~al.}(2014)Okuzono, Otsuru, Tomiku, and
		Okamoto}]{okuzono2014finite}
	\bibinfo{author}{Okuzono, T.}, \bibinfo{author}{Otsuru, T.},
	\bibinfo{author}{Tomiku, R.},  and \bibinfo{author}{Okamoto, N.}
	(\textbf{\bibinfo{year}{2014}}). \enquote{\bibinfo{title}{A finite-element
			method using dispersion reduced spline elements for room acoustics
			simulation}} \bibinfo{journal}{Applied Acoustics} \textbf{79},
	\bibinfo{pages}{1--8}.
	
	\bibitem[{Parsons and Jones(2000)}]{parsons2000acoustic}
	\bibinfo{author}{Parsons, S.},  and \bibinfo{author}{Jones, G.}
	(\textbf{\bibinfo{year}{2000}}). \enquote{\bibinfo{title}{Acoustic
			identification of twelve species of echolocating bat by discriminant function
			analysis and artificial neural networks}} \bibinfo{journal}{Journal of
		experimental biology} \textbf{203}(17), \bibinfo{pages}{2641--2656}.
	
	\bibitem[{Peterson(1986)}]{peterson1986simulating}
	\bibinfo{author}{Peterson, P.~M.} (\textbf{\bibinfo{year}{1986}}).
	\enquote{\bibinfo{title}{Simulating the response of multiple microphones to a
			single acoustic source in a reverberant room}} \bibinfo{journal}{The Journal
		of the Acoustical Society of America} \textbf{80}(5),
	\bibinfo{pages}{1527--1529}.
	
	\bibitem[{Pietrzyk(1998)}]{pietrzyk1998computer}
	\bibinfo{author}{Pietrzyk, A.} (\textbf{\bibinfo{year}{1998}}).
	\enquote{\bibinfo{title}{Computer modeling of the sound field in small
			rooms}} in \emph{\bibinfo{booktitle}{Audio Engineering Society Conference:
			15th International Conference: Audio, Acoustics \& Small Spaces}},
	\bibinfo{organization}{Audio Engineering Society}.
	
	\bibitem[{Prawda \emph{et~al.}(2020)Prawda, Schlecht, V{\"a}lim{\"a}ki
		\emph{et~al.}}]{prawda2020evaluation}
	\bibinfo{author}{Prawda, K.}, \bibinfo{author}{Schlecht, S.~J.},
	\bibinfo{author}{V{\"a}lim{\"a}ki, V.} \emph{et~al.}
	(\textbf{\bibinfo{year}{2020}}). \enquote{\bibinfo{title}{Evaluation of
			reverberation time models with variable acoustics}} in
	\emph{\bibinfo{booktitle}{Sound and Music Computing Conference}}.
	
	\bibitem[{Rathsam and Rafaely(2015)}]{rathsam2015analysis}
	\bibinfo{author}{Rathsam, J.},  and \bibinfo{author}{Rafaely, B.}
	(\textbf{\bibinfo{year}{2015}}). \enquote{\bibinfo{title}{Analysis of
			absorption in situ with a spherical microphone array}}
	\bibinfo{journal}{Applied Acoustics} \textbf{89}, \bibinfo{pages}{273--280}.
	
	\bibitem[{Richard \emph{et~al.}(2017)Richard, Fernandez-Grande, Brunskog, and
		Jeong}]{richard2017estimation}
	\bibinfo{author}{Richard, A.}, \bibinfo{author}{Fernandez-Grande, E.},
	\bibinfo{author}{Brunskog, J.},  and \bibinfo{author}{Jeong, C.-H.}
	(\textbf{\bibinfo{year}{2017}}). \enquote{\bibinfo{title}{Estimation of
			surface impedance at oblique incidence based on sparse array processing}}
	\bibinfo{journal}{The Journal of the Acoustical Society of America}
	\textbf{141}(6), \bibinfo{pages}{4115--4125}.
	
	\bibitem[{Rife and Vanderkooy(1999)}]{Rife1989}
	\bibinfo{author}{Rife, D.},  and \bibinfo{author}{Vanderkooy, J.}
	(\textbf{\bibinfo{year}{1999}}). \enquote{\bibinfo{title}{Transfer-function
			measurement with maximum length sequences}} \bibinfo{journal}{Journal of the
		Audio Engineering Society} \textbf{37}(6), \bibinfo{pages}{419--444}.
	
	\bibitem[{Samarasinghe \emph{et~al.}(2018)Samarasinghe, Abhayapala, Lu, Chen,
		and Dickins}]{samarasinghe2018spherical}
	\bibinfo{author}{Samarasinghe, P.~N.}, \bibinfo{author}{Abhayapala, T.~D.},
	\bibinfo{author}{Lu, Y.}, \bibinfo{author}{Chen, H.},  and
	\bibinfo{author}{Dickins, G.} (\textbf{\bibinfo{year}{2018}}).
	\enquote{\bibinfo{title}{Spherical harmonics based generalized image source
			method for simulating room acoustics}} \bibinfo{journal}{The Journal of the
		Acoustical Society of America} \textbf{144}(3), \bibinfo{pages}{1381--1391}.
	
	\bibitem[{Scheibler \emph{et~al.}(2018)Scheibler, Bezzam, and
		Dokmani{\'c}}]{scheibler2018pyroomacoustics}
	\bibinfo{author}{Scheibler, R.}, \bibinfo{author}{Bezzam, E.},  and
	\bibinfo{author}{Dokmani{\'c}, I.} (\textbf{\bibinfo{year}{2018}}).
	\enquote{\bibinfo{title}{Pyroomacoustics: A python package for audio room
			simulation and array processing algorithms}} in
	\emph{\bibinfo{booktitle}{2018 IEEE International Conference on Acoustics,
			Speech and Signal Processing (ICASSP)}}, \bibinfo{organization}{IEEE}, pp.
	\bibinfo{pages}{351--355}.
	
	\bibitem[{Schimmel \emph{et~al.}(2009)Schimmel, Muller, and
		Dillier}]{schimmel2009fast}
	\bibinfo{author}{Schimmel, S.~M.}, \bibinfo{author}{Muller, M.~F.},  and
	\bibinfo{author}{Dillier, N.} (\textbf{\bibinfo{year}{2009}}).
	\enquote{\bibinfo{title}{A fast and accurate “shoebox” room acoustics
			simulator}} in \emph{\bibinfo{booktitle}{2009 IEEE International Conference
			on Acoustics, Speech and Signal Processing}}, \bibinfo{organization}{IEEE},
	pp. \bibinfo{pages}{241--244}.
	
	\bibitem[{Schr{\"o}der(2011)}]{schroder2011physically}
	\bibinfo{author}{Schr{\"o}der, D.} (\textbf{\bibinfo{year}{2011}}).
	\emph{\bibinfo{title}{Physically based real-time auralization of interactive
			virtual environments}}, \bibinfo{volume}{\textbf{11}}
	(\bibinfo{publisher}{Logos Verlag Berlin GmbH}).
	
	\bibitem[{Schroeder(1965)}]{Schroeder:65}
	\bibinfo{author}{Schroeder, M.~R.} (\textbf{\bibinfo{year}{1965}}).
	\enquote{\bibinfo{title}{New method of measuring reverberation time}}
	\bibinfo{journal}{The Journal of the Acoustical Society of America}
	\textbf{37}, \bibinfo{pages}{409--412}.
	
	\bibitem[{Schroeder(1979)}]{Schroeder1979}
	\bibinfo{author}{Schroeder, M.~R.} (\textbf{\bibinfo{year}{1979}}).
	\enquote{\bibinfo{title}{Integrated-impulse method measuring sound decay
			without using impulses}} \bibinfo{journal}{The Journal of the Acoustical
		Society of America} \textbf{66}(2), \bibinfo{pages}{497--500}.
	
	\bibitem[{Schroeder(1996)}]{Schroeder:96}
	\bibinfo{author}{Schroeder, M.~R.} (\textbf{\bibinfo{year}{1996}}).
	\enquote{\bibinfo{title}{The schroeder frequency revisited}}
	\bibinfo{journal}{The Journal of the Acoustical Society of America}
	\textbf{99}, \bibinfo{pages}{3240--3241}.
	
	\bibitem[{Sides and Mulholland(1971)}]{Sides1971}
	\bibinfo{author}{Sides, D.~J.},  and \bibinfo{author}{Mulholland, K.~A.}
	(\textbf{\bibinfo{year}{1971}}). \enquote{\bibinfo{title}{The variation of
			normal layer impedance with angle of incidence}} \bibinfo{journal}{Journal of
		Sound and Vibration} \textbf{14}, \bibinfo{pages}{139--142}.
	
	\bibitem[{Stan \emph{et~al.}(2002)Stan, Embrechts, and Archambeau}]{Stan2002}
	\bibinfo{author}{Stan, G.-B.}, \bibinfo{author}{Embrechts, J.-J.},  and
	\bibinfo{author}{Archambeau, D.} (\textbf{\bibinfo{year}{2002}}).
	\enquote{\bibinfo{title}{Comparison of different impulse response measurement
			techniques}} \bibinfo{journal}{Journal of the Audio Engineering Society}
	\textbf{50}(4), \bibinfo{pages}{249--262}.
	
	\bibitem[{Suzuki \emph{et~al.}(1995)Suzuki, Asano, Kim, and Sone}]{Suzuki1995}
	\bibinfo{author}{Suzuki, Y.}, \bibinfo{author}{Asano, F.},
	\bibinfo{author}{Kim, H.},  and \bibinfo{author}{Sone, T.}
	(\textbf{\bibinfo{year}{1995}}). \enquote{\bibinfo{title}{An optimum
			computer-generated pulse signal suitable for the measurement of very long
			impulse responses}} \bibinfo{journal}{The Journal of the Acoustical Society
		of America} \textbf{97}(2), \bibinfo{pages}{1119--1123}.
	
	\bibitem[{Tamura(1990)}]{tamura1990spatial}
	\bibinfo{author}{Tamura, M.} (\textbf{\bibinfo{year}{1990}}).
	\enquote{\bibinfo{title}{Spatial fourier transform method of measuring
			reflection coefficients at oblique incidence. i: Theory and numerical
			examples}} \bibinfo{journal}{The Journal of the Acoustical Society of
		America} \textbf{88}(5), \bibinfo{pages}{2259--2264}.
	
	\bibitem[{Torras-Rosell and Jacobsen(2010)}]{Torras2010}
	\bibinfo{author}{Torras-Rosell, A.},  and \bibinfo{author}{Jacobsen, F.}
	(\textbf{\bibinfo{year}{2010}}). \enquote{\bibinfo{title}{Measuring long
			impulse responses with pseudorandom sequences and sweep signals}} in
	\emph{\bibinfo{booktitle}{Internoise}}, \bibinfo{address}{Lisbon, Portugal}.
	
	\bibitem[{Vorl{\"a}nder and Mommertz(2000)}]{vorlander2000definition}
	\bibinfo{author}{Vorl{\"a}nder, M.},  and \bibinfo{author}{Mommertz, E.}
	(\textbf{\bibinfo{year}{2000}}). \enquote{\bibinfo{title}{Definition and
			measurement of random-incidence scattering coefficients}}
	\bibinfo{journal}{Applied acoustics} \textbf{60}(2),
	\bibinfo{pages}{187--199}.
	
	\bibitem[{Wabnitz \emph{et~al.}(2010)Wabnitz, Epain, Jin, and
		Van~Schaik}]{wabnitz2010room}
	\bibinfo{author}{Wabnitz, A.}, \bibinfo{author}{Epain, N.},
	\bibinfo{author}{Jin, C.},  and \bibinfo{author}{Van~Schaik, A.}
	(\textbf{\bibinfo{year}{2010}}). \enquote{\bibinfo{title}{Room acoustics
			simulation for multichannel microphone arrays}} in
	\emph{\bibinfo{booktitle}{Proceedings of the International Symposium on Room
			Acoustics}}, \bibinfo{organization}{Citeseer}, pp. \bibinfo{pages}{1--6}.
	
	\bibitem[{Yu and Kleijn(2020)}]{yu2020room}
	\bibinfo{author}{Yu, W.},  and \bibinfo{author}{Kleijn, W.~B.}
	(\textbf{\bibinfo{year}{2020}}). \enquote{\bibinfo{title}{Room acoustical
			parameter estimation from room impulse responses using deep neural networks}}
	\bibinfo{journal}{IEEE/ACM Transactions on Audio, Speech, and Language
		Processing} \textbf{29}, \bibinfo{pages}{436--447}.
	
	\bibitem[{Yuzawa(1975)}]{Yuzawa1975}
	\bibinfo{author}{Yuzawa, M.} (\textbf{\bibinfo{year}{1975}}).
	\enquote{\bibinfo{title}{A method of obtaining the oblique incident sound
			absorption coefficient through an on-the-spot measurement}}
	\bibinfo{journal}{Applied Acoustics} \textbf{8}, \bibinfo{pages}{27--41}.
	
\end{thebibliography}

\end{document}